\documentclass[10pt,twocolumn,letterpaper]{article}

\usepackage{iccv}
\usepackage{times}
\usepackage{epsfig}
\usepackage{graphicx}
\usepackage{amsmath}
\usepackage{amssymb}
\usepackage{subfigure}
\usepackage{times}
\usepackage{bm}
\usepackage{ulem}
\usepackage{booktabs}
\usepackage{multirow}
\usepackage{footnote}

\usepackage[pagebackref=true,breaklinks=true,letterpaper=true,colorlinks,bookmarks=false]{hyperref}

\iccvfinalcopy 

\def\iccvPaperID{4134} 

\newtheorem{theorem}{Theorem}
\newtheorem{definition}{Definition}

\begin{document}

\title{Unsupervised Learning of Multi-level Structures for Anomaly Detection}

\author{Songmin Dai, Jide Li, Lu Wang, Congcong Zhu, Yifan Wu, Xiaoqiang Li\thanks{Corresponding author}\\
School of Computer Engineering and Science, Shanghai University, China\\
{\tt\small \{laodar,iavtvai,luwang,congcongzhu,VictorWu,xqli\}@shu.edu.cn}
}

\maketitle
\ificcvfinal\thispagestyle{empty}\fi

\begin{abstract}
   The main difficulty in high-dimensional anomaly detection tasks is the lack of anomalous data for training. And simply collecting anomalous data from the real world, common distributions, or the boundary of normal data manifold may face the problem of missing anomaly modes. This paper first introduces a novel method to generate anomalous data by breaking up global structures while preserving local structures of normal data at multiple levels. It can efficiently expose local abnormal structures of various levels. To fully exploit the exposed multi-level abnormal structures, we propose to train multiple level-specific patch-based detectors with contrastive losses. Each detector learns to detect local abnormal structures of corresponding level at all locations and outputs patchwise anomaly scores. By aggregating the outputs of all level-specific detectors, we obtain a model that can detect all potential anomalies. The effectiveness is evaluated on MNIST, CIFAR10, and ImageNet10 dataset, where the results surpass the accuracy of state-of-the-art methods. Qualitative experiments demonstrate our model is robust that it unbiasedly detects all anomaly modes.
\end{abstract}

\section{Introduction}

\label{sec:intro}
	Detecting anomalies or out-of-distribution (OOD) data is very important in different application domains, such as fraud detection for credit cards, defect detection of industrial products, lesion detection based on medical images, security check based on X-ray images, and video monitoring~\cite{ComplentaryGAN, MVTecAD, app1, app2, GANomaly}.
	Although it has been studied in many fields for long, the anomaly detection for image data is still difficult to handle~\cite{LessBiased}.
	Machine learning models like deep neural networks often have a vast hypothesis space and rely on the i.i.d. (independent and identically distributed) assumption. By optimizing the training objectives, many functions can be found with desirable outputs on the in-distribution data but undefined outputs on the OOD data. Similarly, we cannot explicitly constraint an anomaly detector on those anomaly modes not used for training, and their predictions may not be desirable. 
	
	\begin{figure}[!t]
		\includegraphics[width=\linewidth]{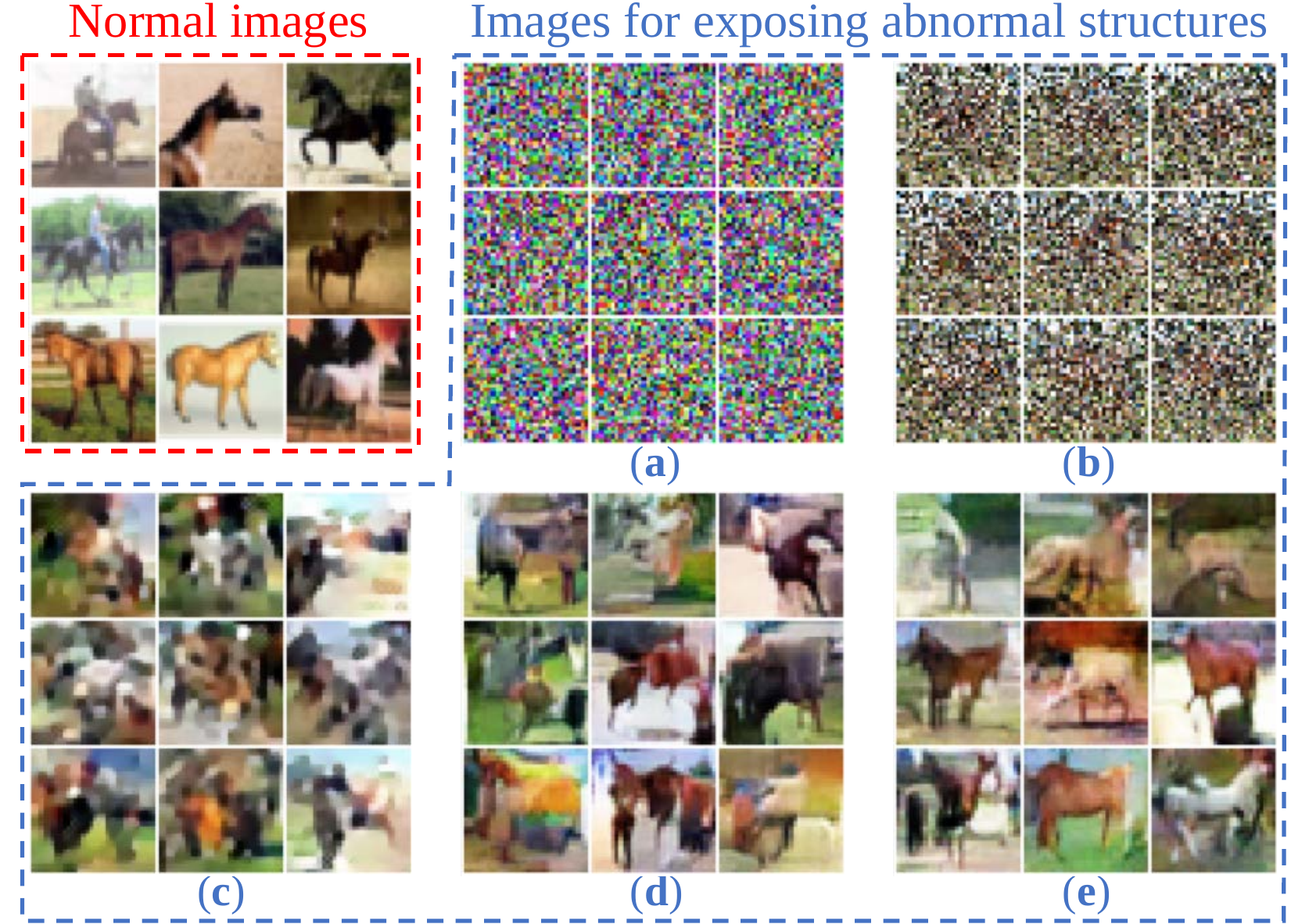}
		\caption{Images used for learning multi-level visual structures of CIFAR10 horse class. (\textbf{a})-(\textbf{e}) are sampled from the MRF approximations (Definition~\ref{definition_MRF}) of level 0, 1, 2, 4, and 8. They recombine the visual elements of normal images (Theorem~\ref{theorem_supp_eq}) and produce abundant local abnormal structures at various levels.} 
		\label{first_page}
	\end{figure}
	
	In low-dimensional tasks, it is easy to generate data that cover all anomaly modes from a noise distribution supported on the whole data space. We can train a binary classifier that learns the statistics of the normal data by contrasting the normal data with the generated noise data, like Noise Contrastive Estimation (NCE) ~\cite{NCE,NCE0}. 
	However, in high-dimensional tasks, there are two problems: (i) Almost all noise samples contain only low-level anomalous features, as shown in Fig.~\ref{first_page}(a). They are inefficient to expose higher-level abnormal structures. (ii) Classifiers may degenerate to learn the statistics of some certain regions. For example, they can successfully discriminate the normal data from the generated data based on only a few pixels in the center of an image, without learning anything about other parts~\cite{DLbook}.

	To address the first problem, we first propose to use images from Markov Random Field (MRF) approximations of multiple levels with suitable gaps to expose local abnormal structures of various levels (as shown in Fig.~\ref{first_page}). We then develop a novel generation method called entropy-regularized PatchGAN to obtain such images. To address the second problem (which also exists in images from MRF approximations), we propose to use multiple level-specific patch-based detectors. We train each level-specific detector to learn patch-wise anomaly scores and calculate the final anomaly score by aggregating the results over all patches and levels. We call the proposed model Multi-Level Structure Anomaly Detection (MLSAD).
	
	MLSAD is a robust model that unbiasedly detects all anomaly modes, since it sees diverse local abnormal structures of various levels and efficiently learns to detect them at any location. Such robustness is very important for practical usage, especially for safety-critical environments~\cite{Review} and tasks where we do not know which type or level of anomalies will appear. MLSAD takes no assumption on the image type, so it is a general method for both objects and textures. It achieves state-of-the-art performance compared with methods requiring neither external OOD dataset nor prior knowledge on the image type.
	\section{Related works}
	Many deep unsupervised anomaly detection approaches have been developed in recent years. Some of them~\cite{PatchSVDD} combine deep networks with traditional methods like DeepSVDD~\cite{DeepSVDD}, which learns a neural network that maps the representations of normal data into a hypersphere with minimal volume. Besides, there are many other types of methods, which will be reviewed in the following.
	
	\textbf{Reconstruction based}.
	Training autoencoders (AEs) \cite{AE2,GANomaly,VAE} on normal data is a popular method for deep unsupervised anomaly detection. A well-trained AE is supposed to produce lower reconstruction errors on the normal data than the anomalous data. However, in practice, it may also reconstruct anomalies very well or even better~\cite{Pidhorskyi2018GenerativePN}. In order to reconstruct only the normal data, AnoGAN~\cite{AnoGAN} learns a generator mapping the latent space to the in-distribution manifold and finds the closest in-distribution sample for a given input by searching in the latent space. OCGAN~\cite{OCGAN} learns a bounded latent space that exclusively generates only in-distribution samples. In \cite{ConAD}, it is argued that AEs tend to regress the conditional mean rather than the actual multi-modal distribution and propose to learn a multi-hypotheses autoencoder that finds the closest reconstruction for the input. GPND~\cite{GPND} trains an adversarial autoencoder and improves the detection performance by combining the reconstruction error with latent likelihood.
	
	\textbf{Confidence based}. These methods~\cite{softmax,Rot,Geom,ConfidentClassifier,ODIN} detect OOD data using the softmax activation statistics of multi-class models trained on normal data, with the assumption that anomalies will tend to have lower classification confidence than the correctly classified normal samples. To obtain classifiers with unlabeled normal data, \cite{Geom} train networks to classify the geometric transformations applied to inputs. This idea is combined with adversarial training to improve both anomaly detection performance and robustness of classifiers in \cite{Rot}. It is also extended to non-image data using random affine transformations by \cite{GOAD}. 
	
	\textbf{Density based}.
	Another appealing idea for unsupervised anomaly detection is estimating the density of the in-distribution via deep generative models. However, recent works \cite{WAIC,DoGMsKnow} show that these models assign higher likelihoods to OOD samples even than in-distribution samples. \cite{complexity} proposes to address it by subtracting the input complexity from the original OOD score. \cite{OE} finds training against an auxiliary dataset of outliers significantly improves their performance. \cite{Ratio} proposes a background contrastive score that captures the significance of the semantics compared with the background model for OOD detection. \cite{GlowFail} shows that flow models tend to learn local pixel correlations rather than semantic structures and propose to modify the architecture of flow coupling layers.
	
	\textbf{Anomaly generation based}.
	A common issue for most previous methods is that there is no direct constraint on those OOD samples, and thus some anomalies may have lower anomaly scores than the normal samples'. \cite{Relu} shows that ReLU type neural networks always have high confidence predictions on samples far away from the training data and propose to use uniform noise or adversarial noise samples to enforce low confidence on the OOD data. But noise distributions will be less efficient when applied to high dimensional data. Some works~\cite{ALOCC,ASG,OODGen} proposed to generate anomalous samples near the boundary of the in-distribution and thus obtain hard samples that are more efficient for learning a tight decision boundary. However, there is still no guarantee for handling all anomalies (e.g., the anomalies far away from the in-distribution data).
	
	Unlike Geom~\cite{Geom}, Rot~\cite{Rot} and the recent contrastive-learning-based model CSI~\cite{CSI}, MLSAD does not rely on discriminating geometrically transformed images or contrasting with (manually selected) dataset-dependent shifted instances. Therefore it is less limited by the image type of the in-distribution. Compared with previous anomaly generation based methods, we explicitly take the multi-level anomalous structures into consideration and thus design a multiple level-specific patch-based detection framework to overcome the problem of missing anomaly modes.
	\section{Approaches}
	\subsection{Exposing Multi-level Abnormal Structures}
	The main insight into exposing multi-level abnormal structures is breaking up the global structures while preserving local structures at various levels. 
	Take text data as an example. We can expose spelling errors and grammar errors by recombining the elements of correct sentences at letter level and word level, respectively:\\
	\centerline{\textbf{Letter level}: rdoonyaa. dd otca e,d e lregg  gadfo}
	\centerline{\textbf{  Word level}: green add lead good, go for day cat}
	Letter-level recombinations can successfully avoid illegal characters and produce spelling errors, but they cannot efficiently expose errors coming from wrong combinations of correct words (like grammar errors), because it is hard to produce a text sequence with each word correctly spelled. Instead, word-level recombinations can successfully avoid spelling errors, focusing more on producing advanced errors, and thus more efficient to expose grammar errors. In the following, we formalize the idea of exposing multi-level abnormal structures by breaking up global structures while preserving local structures at various levels and generalize it to the image data via MRF approximations.
	
	\subsubsection{Markov Random Field Approximation.}
	Let $X$ be the image variable, $c\in C^w$ be an image patch of $w\!\times\!w$ pixels, where $C^w$ denotes the set of all such patches. We use $X_c$ to denote the patch variable on $c$. We define the MRF approximations as following:
	\begin{definition}
		\label{definition_MRF}
		\textbf{(MRF approximations)}
		A distribution $q_{\bm{X}}$ is called the $w${\rm th}-level MRF approximation of a given image distribution $p_{\bm{X}}$, if for each patch $c$, the margial distribution of $X_c$ from $q_{\bm{X}}$ is identical to the one from $p_{\bm{X}}$: $\forall c\in C^w, q_{\bm{X}_c}(\bm{x}_c) = p_{\bm{X}_c}(\bm{x}_c)$, and the density function of $q_{\bm{X}}$ can be factored as the product of functions of patch variables: $q_{\bm{X}}(\bm{x})=\prod_{c\in C^w} \phi_c(\bm{x}_c)$. Particularly, the uniform distribution is defined to be the $0$th-level MRF approximation of any image distribution.
	\end{definition}
	We will use $p_{\bm{X}}$ to denote the distribution of normal images and $q^w_{\bm{X}}$ to denote the $w$th-level MRF approximation of it. By Definition~\ref{definition_MRF}, we have $q_{\bm{X}_c}^w(\bm{x}_c)=p_{\bm{X}_c}(\bm{x}_c)$ for any patch $c\in C^w$. It means images from $q^w_{\bm{X}}$ are indistinguishable from normal images via any window of $w$$\times$$w$ pixels, and the local structures of normal images are well kept. 
	Besides, there are no long-range interactions among patches since $q^w_{\bm{X}}$ can be factored as $\prod_{c\in C^w} \phi_c(\bm{x}_c)$. It breaks the global structures as much as possible and makes the local structures recombined more randomly. Moreover, we conclude that all recombinations can be sampled from $q^w_{\bm{X}}$ with nonzero probability by the following theorem. 
	\begin{theorem}
		\label{theorem_supp_eq}
		Let $S_w=\{\bm{x}|\forall c\in C^w,p_{\bm{X}_c}(\bm{x}_c)>0\}$, we have $supp(q^w_{\bm{X}})=S_w$. In other words, the support set of $q^w_{\bm{X}}$ is identical to the set of all images where each patch of $w\times w$ pixels is normal.
	\end{theorem}
	All proofs are provided in Appendix A.\\
	 However, just as letter-level recombinations of sentences can efficiently expose the word-level errors but few sentence-level errors, $q^w_{\bm{X}}$ will be only efficient to expose the abnormal structures of levels slightly higher than $w$. This paper assumes that we can efficiently expose all local abnormal structures of all levels by MRF approximations of multiple levels with suitable gaps.
	
	\begin{figure*}[!t]
		\includegraphics[width=\linewidth]{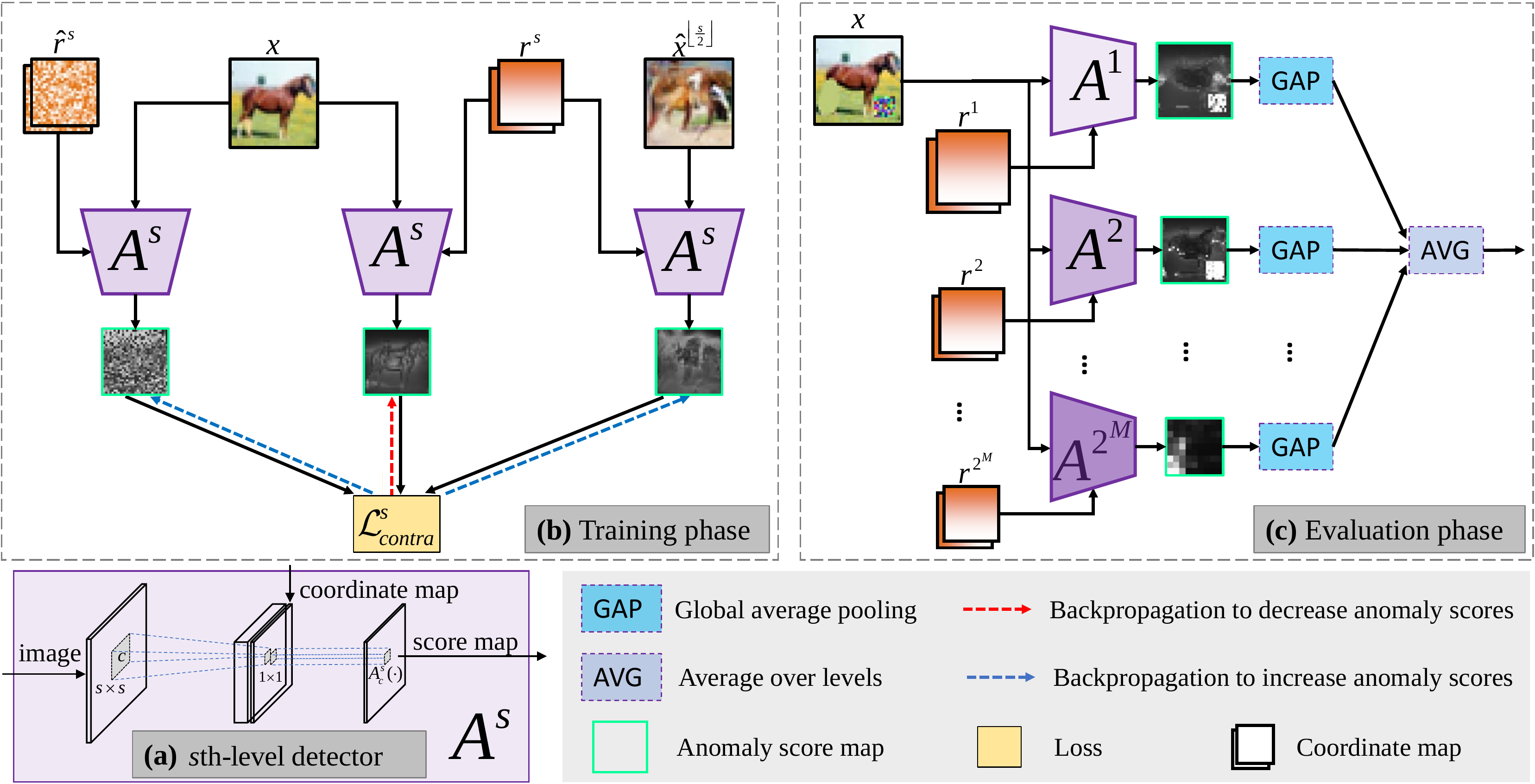}
		\caption{Overview of \textit{Multi-Level Structure Anomaly Detection} (MLSAD). Our model contains $M$$+$$1$ level-specific patch-based detectors $\{A^{s}|s\!=\!1,2,4,...,2^M\}$. \textbf{(a)} Each $s$th-level detector $A^{s}$ outputs a position-aware anomaly score for each patch of $s\!\times\! s$ pixels. \textbf{(b)} $A^s$ is trained to detect position-dependent local abnormalities between level $\lfloor\frac{s}{2}\rfloor\!+\!1$ and $s$, by contrasting the normal pair $(\bm{x},\bm{r}^s)$ with two anomalous pairs, $(\hat{\bm{x}}^{\lfloor\frac{s}{2}\rfloor},\bm{r}^s)$ and $(\bm{x},\hat{\bm{r}}^s)$, as Eq.~\ref{loss_function}, where $\bm{x}$ is a normal image, $\hat{\bm{x}}^{\lfloor\frac{s}{2}\rfloor}$ is a fake image sampled from $\lfloor\frac{s}{2}\rfloor$th-level MRF approximation, $\bm{r}^s$ and $\hat{\bm{r}}^s$ are a true coordinate map and a fake coordinate map, their shapes are compatible with $A^s$. \textbf{(c)} During evaluation phase, an overall anomaly score for the test image $\bm{x}$ is computed by simply averaging the outputs of all level-specific detectors.}
		\label{mainframe}
	\end{figure*}
	
	\subsubsection{Entropy-regularized PatchGAN.}
	By definition~\ref{definition_MRF}, the $0$th-level MRF approximation $q^0_{\bm{X}}$ can be obtained by the uniform distribution, and the $1$st-level MRF approximation $q^1_{\bm{X}}$ can be easily obtained by shuffling pixels of the normal images along the batch axis. To generate the higher-level ($w\geq 2$) MRF approximations $q^w_{\bm{X}}$, we propose a Generative Adversarial Networks (GAN~\cite{GAN}) based approach. Like previous works~\cite{pix2pix,SinGAN,PSGAN} that generate images with all locals realistic or tile image patches seamlessly, we train the generator $G$ that maps a noise $\bm{z}$ to an image together with a patch-based discriminator $D$. $D$ is implemented with a Fully Convolutional Network (FCN~\cite{FCN}) with stride 1. For the $w$th-level MRF approximation, $D$ has a receptive filed size of $w\times w$ pixels, and outputs a score $D_c$ for each patch $c\in C^w$. The adversarial losses are averaged over all patches for both $D$ and $G$:
	\begin{equation*}
		\begin{aligned}
			\mathcal{L}_{adv}(D)=-\frac{1}{|C^w|}\sum_c [\mathbb {E}_{\bm{x}}&\log D_c(\bm{x})+\\
			\mathbb {E}_{\bm{z}}&\log(1\!-\!D_c(G(\bm{z})))],\\
			\mathcal{L}_{adv}(G)=-\frac{1}{|C^w|}\sum_c [\mathbb {E}_{\bm{z}}&\log D_c(G(\bm{z}))].
		\end{aligned}
	\end{equation*}
	We feed coordinate information into $D$ like~\cite{CoordConv}, since $p_{\bm{X}_c}$ may be varied with patch $c$. 
	Using the adversarial losses alone can ensure the local visual structures are well kept, while the global structures may not be thoroughly broken, since there are often some inherent global structures caused by the inductive bias of $G$.
	\begin{theorem}
		\label{theorem_max_entropy}
		The maximum entropy distribution satisfying $\forall c\!\in\!C^w,\quad q_{\bm{X}_c}(\bm{x}_c)\!=\! p_{\bm{X}_c}(\bm{x}_c)$ can be factored as $\prod_{c\in C^w} \phi_c(\bm{x}_c)$.
	\end{theorem}
	By Theorem 2 and Definition 1, we can know that the maximum entropy distribution preserving local structures in windows of $w\times w$ pixels is exactly the $w$th-level MRF approximation of $p_{\bm{X}}$. So we turn to maximize the entropy of the outputs of the generator while training it with a GAN loss. Similar to \cite{EBGAN}, we adopt the recently proposed mutual information maximization technique. It relies on two facts: (i) The entropy of the generated distribution $H(G(Z))$ is identical to the mutual information between $Z$ and $G(Z)$, because $I(G(Z),Z)=H(G(Z))-H(G(Z)|Z)$ and $H(G(Z)|Z)=0$ since $G$ is deterministic. (ii) The mutual information between $X$ and $Z$ can be estimated by the neural mutual information measure~\cite{MINE}: 
	\begin{equation}
	\label{I_JS}
	\begin{aligned}
	\mathcal{I}_{\Theta}(X,Z)=\sup_{T\in \mathcal{T}_{_{\Theta}}}[&\mathbb{ E }_{p(X,Z)}[T(X,Z)]-\\
	&\log(\mathbb{ E }_{p(X)p(Z)}[e^{T(X,Z)}])].
	\end{aligned}{\tiny }
	\end{equation}
	The supremum is approximated using gradient descent on the parameters of a statistics network $T$ (also updated in synchronization with $G$). Thus the total loss for $G$ becomes:
	\begin{equation}
	\mathcal{L}_{total}(G) = \mathcal{L}_{adv}(G) - \beta\mathcal{I}_{\Theta}(G(Z),Z).
	\end{equation}
	We can interpret the first term as an objective for preserving local structures and the second term for breaking global structures. These two terms work together to create diverse and seamless images. They will be used to expose local abnormal structures of levels slightly higher than $w$. We will refer to this generation approach as entropy-regularized PatchGAN in the following.
	\begin{figure*}[!htbp]
		\includegraphics[width=\linewidth]{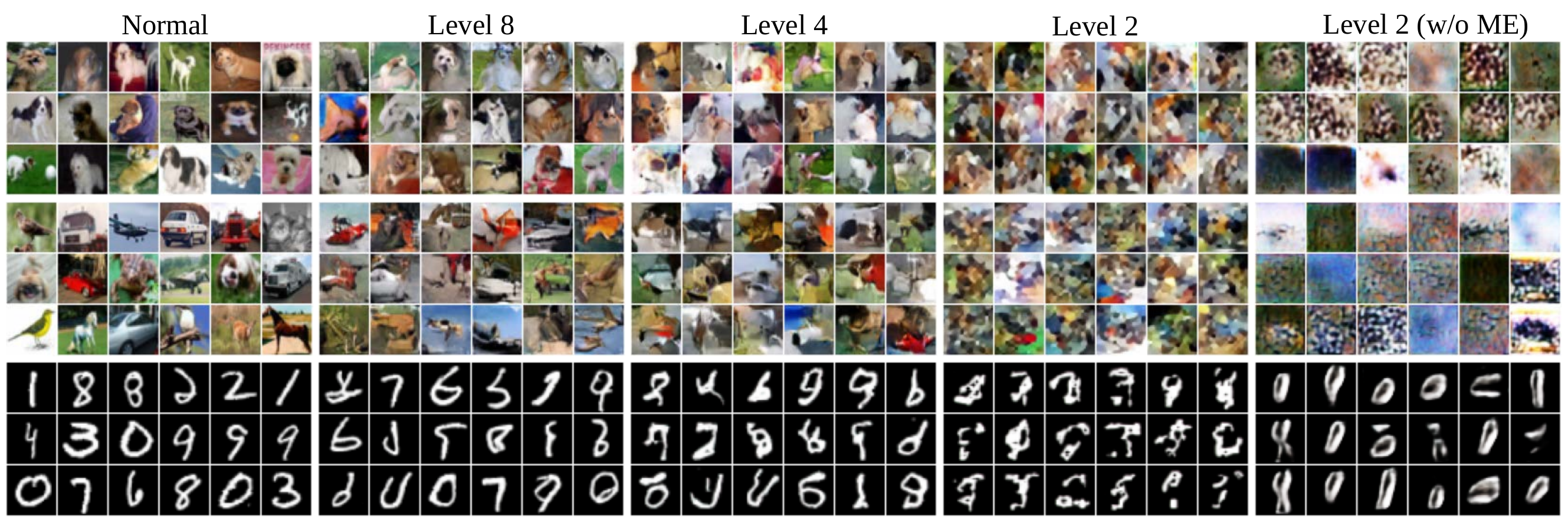}
		\caption{Generation results of entropy-regularized PatchGAN. The leftmost column block shows the images of the normal classes used for training. The next three column blocks show the generations of levels 8, 4 and 2, by entropy-regularized PatchGANs. For comparison, the rightmost column block shows the generations of level 2 without maximum entropy regularization.}
		\label{generations}
	\end{figure*}
	\subsection{Multi-Level Structure Anomaly Detection}
	Without loss of generality, we assume the image size is $2^W\!\times2^W$ pixels, and that local abnormal structures of all levels can be efficiently exposed by MRF approximations with suitable level gaps $\{q^{\lfloor \frac{s}{2}\rfloor}_{\bm{X}}|s\!=\!1,2,4,...,2^W\}$. However, we remark that any MRF approximation cannot efficiently expose any image with anomalous structures of multiple levels, and that images from MRF approximations (especially for low levels) often show abnormal cues at almost all locations. An image-level detector trained to discriminate between such images and the normal images may degenerate to learn the statistics of a few regions and cannot generalize to all anomalous images. To fully exploit the exposed local abnormal structures, learn the visual structures of all levels and at all locations, and handle various anomaly modes, we propose to train multiple level-specific patch-based detectors and aggregate their outputs for a final prediction, as shown in Fig.~\ref{mainframe}. We name the proposed method as Multi-Level Structure Anomaly Detection (MLSAD). 
	\subsubsection{The $\bm{s}$th-level detector} The MLSAD model consists of $M+1$ level-specific detectors, $\{A^{s}|s\!=\!1,2,4,...,2^M\}$, where $A^s$ is the $s$th-level detector and $M\leq W$. $A^s$ outputs an anomaly score $A^s_c(\cdot)\in[0,1]$ for each patch $c$ in $C^s$ by an FCN (with stride of $1\!\times\! 1$ and receptive field size of $s\!\times\!s$ pixels). A coordinate map is used as the auxiliary input to give $A^s_c$ the potential to learn position-dependent statistics, see Fig.~\ref{mainframe}(a).
	\subsubsection{Training Phase} 
	For each level-specific detector $A^s$, we train it to detect the position-dependent patch abnormalities between level $\lfloor\frac{s}{2}\rfloor\!+\!1$ and $s$. The basic idea is to use the fake image $\hat{\bm{x}}^{\lfloor \frac{s}{2} \rfloor}$ from $\lfloor \frac{s}{2} \rfloor$th-level MRF approximation for contrast, and force $A^s$ to discriminate the normal image $\bm{x}$ and $\hat{\bm{x}}^{\lfloor \frac{s}{2} \rfloor}$ in each window of $s\!\times\!s$ pixels. However, we observed $A^s$ tend to focus on learning the common statistics shared among patches and utilize very little position information (See Appendix B). In the perspective of the output neurons of $A^s$, they compute the anomaly score for each patch using the same convolution kernels, and struggle to learn a position-aware score by modeling the dependence between the patch content and the coordinates. Such dependence can also be viewed as some kind of high-level structure and may be hard to learn. To alleviate this concern, we propose to use the normal image $\bm{x}$ paired with a fake coordinate map $\hat{\bm{r}}^s$ as additional contrastive data. These contrastive data provide samples breaking the dependence between position and local content.
	
	The overall training method is shown in Fig.~\ref{mainframe}(b). The normal image paired with true coordinate map $(\bm{x},\bm{r}^s)$ is designed as the normal pair, while both the structure anomaly pair $(\hat{\bm{x}}^{\lfloor \frac{s}{2} \rfloor},\bm{r}^s)$ and the position anomaly pair $(\bm{x},\hat{\bm{r}}^s)$ are designed as the anomalous pairs. We train $A^s$ to (patchwise) classify the anomalous pairs as positive class and the normal pair as negative class via the following loss:
	\begin{equation}\label{loss_function}
	\begin{aligned}
	\mathcal{L}^{s}_{contra}=-\frac{1}{|C^s|}\sum_c\Big[&\mathbb{E}_{(\bm{x},\bm{r}^s)}\log (1-A_c^s(\bm{x}, \bm{r}^s))\\
	+\alpha_{1}&\mathbb{E}_{(\bm{x}, \hat{\bm{r}}^s)}\log A_{c}^{s}(\bm{x}, \hat{\bm{r}}^{s})\\
	+\alpha_{2}&\mathbb{E}_{(\hat{\bm{x}}^{\lfloor \frac{s}{2} \rfloor},\bm{r}^s)}\log A_{c}^{s}(\hat{\bm{x}}^{\lfloor \frac{s}{2}\rfloor}, \bm{r}^{s})\Big],
	\end{aligned}
	\end{equation}
	where $\alpha_{1},\alpha_{2}\in [0,1]$. The elements of $\bm{r}^s$ are assigned with coordinate values of its corresponding position. Each $\hat{\bm{r}}^s$ is filled with randomly sampled values in coordinate range. We feed the coordinate maps into $A^s$ at a intermediate layer, whose following layers have strides of $1\times1$. This prevents $A^s$ from simply discriminating $(\bm{x},\bm{r})$ from $(\bm{x},\hat{\bm{r}}^s)$ via the local discontinuities in $\hat{\bm{r}}^s$.
	\subsubsection{Evaluation Phase}
	By setting $M$$=$$W$, $\bigcup_{s=1,2,4,...,2^M}\{\lfloor\frac{s}{2}\rfloor\!+\!1,\lfloor\frac{s}{2}\rfloor\!+\!2,\lfloor\frac{s}{2}\rfloor\!+\!3,...,s\}$ can cover all possible levels. Then every anomalous image will trigger at least one level-specific detector with a high response at some location, since every anomalous image contains at least one abnormal structure of some level. Intuitively, we can use max pooling to aggregate the outputs of all levels and locations and obtain a final score that probably detects all possible anomalies. However, in practice, max pooling relies too much on the score of the most anomalous region, which may result in a false positive detection when the test normal image is slightly corrupted. We notice that average pooling and softmax pooling tend to work better ~\cite{Pooling}. Here, we simply perform average pooling over locations and levels successively:
	\begin{equation}
	\label{eq_eval_phase}
	A(\bm{x})=\frac{1}{M+1}\sum_{s}\left[\frac{1}{|C^s|}\sum_{c}A_{c}^{s}(\bm{x}, \bm{r}^{s})\right].
	\end{equation}
	In practice, the detector of very high levels may not be necessary, and setting $M<W$ may also work well, since the structures below a certain level are often enough to capture the most information of normal images~\cite{AmbientGAN,BagNet}, especially for texture-like images.
	\begin{table*}[!h]
		\setlength{\belowcaptionskip}{-0.2cm}
		\setlength{\abovecaptionskip}{0.0cm}
		\centering
		\caption{Anomaly detection results on CIFAR10 dataset using AUROC metric.}
		\resizebox{\textwidth}{!}{
			\begin{tabular}{llllllllllll}
				\toprule
				&	plane&	car&	bird&	cat&	deer&	dog&	frog&	horse&	ship&	truck&mean\\
				\midrule
				\footnotesize\textbf{Limited by image type}\\
				Geom\cite{Geom} & 0.747 & 0.957&0.781&0.724&0.878&0.878&0.834&0.955&0.933&0.913&0.8600\\
				Rot\cite{Rot}& 0.783&0.943&0.862&0.808&0.894&0.890&0.889&0.951&0.923&0.897&0.8840\\
				CSI\cite{CSI} & 0.899&0.991&0.931&0.864&0.939&0.932&0.951&0.987&0.979&0.955&0.9428\\
				\footnotesize\textbf{Not limited by image type}\\
				OCSVM~\cite{OCSVM}&	0.630&	0.440&	0.649&	0.487&	0.735&	0.500&	0.725&	0.533&	0.649&	0.508& 0.5856\\
				KDE~\cite{KDE}& 0.658&0.520&0.657&0.497&0.727&0.496&0.758&0.564&0.680&0.640&0.6097\\
				AnoGAN~\cite{AnoGAN}&	0.671&	0.547&	0.529&	0.545&	0.651&	0.603&	0.585&	0.625&	0.758&	0.665&0.6179\\
				DeepSVDD~\cite{DeepSVDD}&	0.617&	0.659&	0.508&	0.591&	0.609&	0.657&	0.677&	0.673&	0.759&	0.731&0.6481\\
				OCGAN~\cite{OCGAN}&	0.757&	0.531&	0.640&	0.620&	0.723&	0.620&	0.723&	0.575&	0.820&	0.554&0.6566\\
				DROCC~\cite{DROCC}& 0.817& 0.767 & 0.667 & 0.671 & 0.736 & 0.744 & 0.744 & 0.714 & 0.800 & 0.762 & 0.7422\\
				MSLAD (Ours)&\textbf{0.819}&\textbf{0.937}&\textbf{0.746}&\textbf{0.725}&\textbf{0.804}&\textbf{0.812} &\textbf{0.905}&\textbf{0.925}&\textbf{0.884}&\textbf{0.907}&\textbf{0.8459}\\
				\bottomrule							
		\end{tabular}}
		\label{AUC_CIFAR10}
	    \end{table*}
	    
		\begin{table*}[!h]
		\setlength{\belowcaptionskip}{-0.2cm}
		\setlength{\abovecaptionskip}{0.0cm}
		\centering
		\caption{Anomaly detection results on ImageNet10 dataset using AUROC metric. Where $0\!\sim\!9$ denote image classes Tench, English Springer, Cassette Player, Chainsaw, Church, French Horn, Garbage Truck, Gas Pump, Golf Ball and  Parachute.}
		\resizebox{\textwidth}{!}{
			\begin{tabular}{llllllllllll}
				\toprule
				& 0 & 1 & 2 & 3 & 4 & 5 & 6 & 7 & 8 & 9 &mean\\
				\midrule
				Neareast Neighbor&	0.656&	0.564&	0.477&	0.452&	0.614&	0.505&	0.542&	0.474&	0.704&	0.759& 0.5746\\
				DeepSVDD~\cite{DeepSVDD}& 0.651&0.665&0.605&0.594&0.563&0.531&0.622&0.567&0.722&0.814&0.6333\\
				DROCC~\cite{DROCC}&	0.702&	0.705&	0.712&	\textbf{0.686}&	0.675&	\textbf{0.770}&	0.691&	\textbf{0.699}&	0.707&	\textbf{0.935}&0.7283\\
				MLSAD (Ours)& \textbf{0.879}&\textbf{0.843}&\textbf{0.829}&0.642&\textbf{0.841}&0.725&\textbf{0.867}&0.664&\textbf{0.729}&0.801&\textbf{0.7820}\\
				\bottomrule							
		\end{tabular}}
		\label{AUC_ImageNet10}
	\end{table*}
	
	\begin{table*}[!h]
		\setlength{\abovecaptionskip}{0.0cm}
		\setlength{\belowcaptionskip}{-0.2cm}
		\centering
		\caption{Anomaly detection results on MNIST dataset using AUROC metric. The results are averaged over all digit classes.}
		\setlength{\tabcolsep}{0.5mm}{
		\resizebox{\textwidth}{!}{
			\begin{tabular}{l|l|l|l|l|l|l|l}
				\toprule
				KDE~\cite{KDE}&AnoGAN~\cite{AnoGAN}&DeepSVDD~\cite{DeepSVDD} &OCSVM~\cite{OCSVM}&AND~\cite{AND}&OCGAN~\cite{OCGAN}&$\text{MLSAD}_{\alpha_1=0}$&MLSAD (Ours)\\
				\midrule
				 0.8143\quad\quad\quad\quad&0.9127\quad\quad\quad\quad&0.9480\quad\quad\quad\quad&0.9513\quad\quad\quad\quad&0.9671\quad\quad\quad\quad&0.9750\quad\quad\quad\quad&0.9139\quad\quad\quad\quad&\textbf{0.9834}\quad\quad\quad\\
				\bottomrule
		\end{tabular}}}
		\label{AUC_MNIST}
	\end{table*}
	
	\section{Experiments}
	In this section, we will first demonstrate the generation results of entropy-regularized PatchGAN, then report the anomaly detection results on various datasets, and finally investigate the robustness of MLSAD and existing methods. Meanwhile, ablation studies on each component are performed. The details in the following experiments can be found in Appendix C.
	
	\subsection{Anomaly Generation}
	We test entropy-regularized PatchGAN on three normal image sets: MNIST, CIFAR10, and CIFAR10 dog images. All images are preprocessed into a shape of $32\!\times\!32\!\times\!3$ by resizing and color converting if necessary. We generate MRF approximations of levels 2, 4, and 8 for each normal set. For comparison, we also train generators without the maximum entropy regularization for level 2.
	
	\textbf{Results.}
	As shown in Fig.~\ref{generations}, the generations of entropy-regularized PatchGAN at high levels tend to show recognizable patterns and semantic-level anomalous structures. The generations at low levels tend to be more disordered and show low-level anomalous cues. Diverse and multi-level local abnormal structures at various locations can be observed. By contrast, the generations without maximum entropy regularization tend to contain simple global patterns and expose much fewer recombination modes.
	\subsection{Anomaly Detection}
	\begin{table*}[htbp]
		\caption{An ablation study of different levels. We choose frog, dog, and horse class in CIFAR10 as the normal class, various types of other image sets as anomalous class: rCIFAR10 (the rest classes in CIFAR10), CIFAR100, SVHN, MNIST, and Noise (uniform).}
		\centering
		\resizebox{\textwidth}{!}{
			\setlength{\tabcolsep}{1.5mm}{
				\begin{tabular}{cccccccccc}
					\midrule
					\multicolumn{1}{c}{\multirow{2}[4]{*}{$A^1$}} & \multicolumn{1}{c}{\multirow{2}[4]{*}{$A^2$}} & \multicolumn{1}{c}{\multirow{2}[4]{*}{$A^4$}} & \multicolumn{1}{c}{\multirow{2}[4]{*}{$A^8$}} & \multicolumn{1}{c}{\multirow{2}[4]{*}{$A^{16}$}} & \multicolumn{5}{c}{AUROC (frog/dog/horse)} \\
					\cmidrule{6-10}          &       &       &       &       & rCIFAR10 & CIFAR100 & SVHN & MNIST & Noise \\
					\midrule
					\midrule
					$\checkmark$ &       &       &       &       &0.730/0.645/0.765 & 0.712/0.681/0.792 & 0.696/0.592/0.807 & 0.487/0.345/0.836 & \textbf{1.000}/\textbf{1.000}/\textbf{1.000}  \\
					& $\checkmark$ &       &       &       & 0.785/0.741/0.804 & 0.737/0.749/0.816 & 0.878/0.787/0.859 & 0.992/0.787/0.900 & \textbf{1.000}/\textbf{1.000}/\textbf{1.000} \\
					&       & $\checkmark$ &       &       & 0.766/0.804/0.851 & 0.768/0.811/0.879& 0.896/0.877/0.786 & 0.977/\textbf{1.000}/0.977 & 0.674/0.974/0.985 \\
					&       &       & $\checkmark$ &       & 0.894/\textbf{0.821}/0.917 & 0.890/\textbf{0.838}/0.928 & 0.950/\textbf{0.914}/\textbf{0.944} & \textbf{1.000}/\textbf{1.000}/\textbf{1.000} & 0.962/0.984/0.999 \\
					&       &       &       & $\checkmark$ & 0.878/0.780/0.904 & 0.880/0.805/0.909 & 0.946/0.806/0.932 & 0.992/\textbf{1.000}/0.928 & 0.864/0.928/0.939 \\
					\midrule
					$\checkmark$ & $\checkmark$ & $\checkmark$ & $\checkmark$ & $\checkmark$ & \textbf{0.905}/0.812/\textbf{0.925} & \textbf{0.901}/0.835/\textbf{0.936} & \textbf{0.961}/0.863/0.943 & 0.997/\textbf{1.000}/0.992 & \textbf{1.000}/\textbf{1.000}/\textbf{1.000}  \\
					\bottomrule
		\end{tabular}}}%
		\label{tab:level_ablation}%
	\end{table*}
	
	\begin{figure*}[!htbp]
	\centering
	\includegraphics[width=\linewidth]{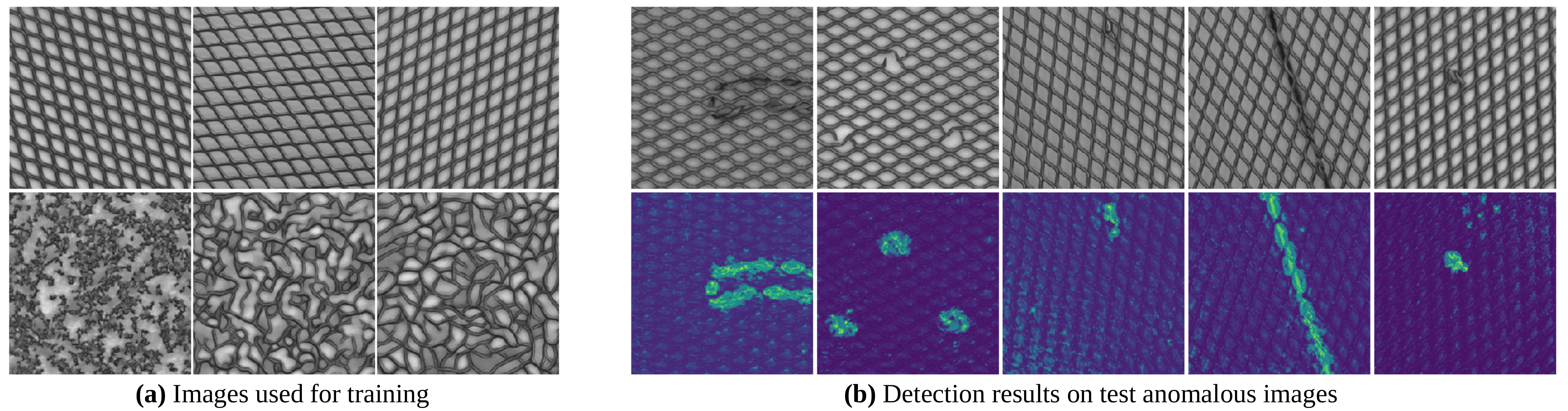}
	\caption{Qualitative results on the 'grid' textures in MVTec AD. (\textbf{a}) Training normal images (top) and multi-level anomalous images $\hat{\bm{x}}^2, \hat{\bm{x}}^4$ and $\hat{\bm{x}}^8$ (bottom). Where $\hat{\bm{x}}^0$ and $\hat{\bm{x}}^1$ are not shown. (\textbf{b}) Test anomalous images and the aggregated pixel-level anomaly scores.}
	\label{texture}
    \end{figure*}

	We evaluate the proposed MSLAD on the handwritten symbol, natural object, and industrial texture images. More specifically, the datasets used are MNIST, CIFAR10, ImageNet10~\cite{ImageNet10} and textures in MVTec AD~\cite{MVTecAD}. For MNIST, CIFAR10 and ImageNet10, we preprocess the images into $32\!\times\!32\!\times\!3$. As the widely used evaluation methodology in prior works~\cite{DROCC,DeepSVDD,OCGAN}, we simulate the unsupervised anomaly detection setting by choosing one out of ten classes as the normal class and the rest as the anomalous class. The original training split of the known class is used for training/validation, and the testing split of all classes is used for testing. We compare MLSAD with deep unsupervised anomaly detection methods using the metric of AUROC (Area Under the Curve of Receiver Operating Characteristics curve). The MLSAD here consists of five level-specific detectors, $\{A^1,A^2,A^4,A^8,A^{16}\}$, where the image-level detector $A^{32}$ is not used since we find it contributes little to the final AUROC. For textures in MVTec AD, we preprocess the images into $256\!\times\!256$ pixels, and likely, we use only the first five level-specific detectors, since we find $\hat{\bm{x}}^{16}$ can already capture almost all the structures of normal images. We train generators with small batch sizes and produce anomalous images with the size of $256\!\times\!256$, while level-specific detectors are trained with randomly cropped patches (size of $32\!\times\!32$) to avoid running out of GPU memory. We obtain the pixel-level anomaly detection results by resizing, aligning, and averaging the output of $A^1,A^2,A^4,A^8$, and $A^{16}$. We simply set $\alpha_1=0.6$ and $\alpha_2=0.4$ for all datasets, despite that we have prior knowledge that the position anomaly pairs will have no contribution to the results of texture images.
	
	\textbf{Results.}
	 Table~\ref{AUC_CIFAR10}, \ref{AUC_ImageNet10} and \ref{AUC_MNIST} present the results on object datasets. Our MLSAD model reaches a better average AUROC value than other unsupervised methods (not limited by image type, e.g. image with rotation-invariant content) on MNIST, CIFAR10, and ImageNet10 dataset, and obtains an improvement on overall CIFAR10 classes compared with the state-of-the-art methods. The results for textures in MVTec AD are shown in Fig.~\ref{texture} and Appendix D, which demonstrate the feasibility of MLSAD for handling translation-invariant and rotation-invariant images. By simply upsampling and averaging the multi-level outputs, MLSAD achieves satisfactory results on pixel-level anomaly localization for textures.
	
	\textbf{Ablation study.} We test the effectiveness of position anomaly pair in the training phase (Eq.~\ref{loss_function}) by setting $\alpha_1\!=\!0$ (and $\alpha_2\!=\!1$). Mean AUC results on MNIST dataset are also shown in Table~\ref{AUC_MNIST}. Like in the texture results, we find the structure anomaly pair working alone can achieve satisfactory performance, while the position anomaly pair indeed further improves the anomaly detection performance (for object images). We then test the effectiveness of the multi-level aggregation used in the prediction phase (Eq.~\ref{eq_eval_phase}). We present the results in table~\ref{tab:level_ablation}. We can observe that the low-level detectors (e.g., $A^1$) have perfect detection accuracy on the noise data but low detection accuracy on more complex anomalous image sets. By contrast, the high-level detectors (e.g., $A^{16}$) have a much better performance on those complex anomalous image sets but poor performance on the noise data. The multi-level aggregation can achieve the best (or close to the best) performance for all anomalous image sets. Thus it is a simple but effective way to build an unbiased model that detects various types of anomalies.
	\begin{figure*}[!t]
		\centering
		\includegraphics[width=0.95\linewidth]{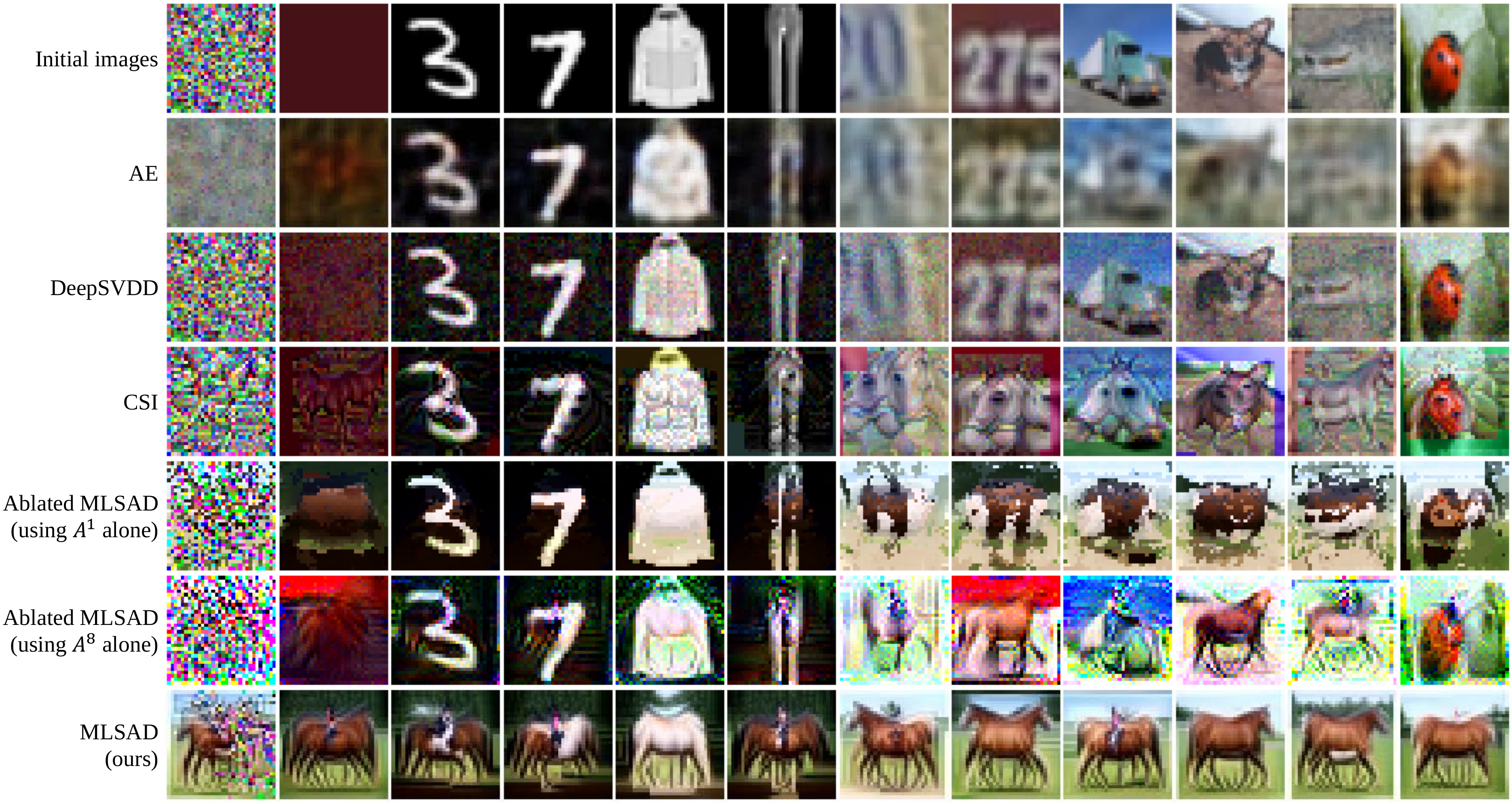}
		\caption{Images found with lower anomaly scores than $95\%$ test normal images' via gradient descend\protect\footnotemark. The detectors used here are trained on CIFAR10 horse images. The top line shows the initial images of the optimizations. The following lines are the images found in various anomaly detection models. The images found in MLSAD have the smallest perceptual differences from the real horse images. We remark that better results for MLSAD may be obtained by using $A^{32}$ and softmax pooling aggregation. See Appendix E for additional results.}
		\label{fooling}
	\end{figure*}
	\subsection{Investigating the robustness.}
	Although previous results have demonstrated the effectiveness of MLSAD to some degree, whether it is robust enough to detect all potential anomalies is still unclear.
	Quantitative evaluation using available anomaly datasets is always biased~\cite{LessBiased} or even misleading~\cite{Review}, since we cannot collect all representative anomalies for a test. To probe the behavior of the MLSAD model on samples outside the normal image distribution, we search in the image space and try to find images with lower anomaly scores than $95\%$ of the test normal images'. We then visualize the found images and check if they do look like the normal images. Similar to prior works~\cite{NNFooled,FirstRobust}, we find such images via gradient descent on the calculated anomaly score. To search as broadly as possible, we start the search from various types of images outside the in-distribution and use a variant of the pull-away term (PT) \cite{PT} as a repelling regularizer:
	\begin{equation}
	L_{PT}(\bm{X_B}) = -\frac{1}{N(N-1)}\sum_{i}{\sum_{j\neq i}{ \|\bm{x}_i-\bm{x}_j\|_1}},
	\label{loos_PT}
	\end{equation}
	where $\bm{X_B}\!=\!\{\bm{x}_1,\bm{x}_2,...,\bm{x}_N\}$ denotes a batch of images being optimized. For comparison, we also perform the search on ablated MLSADs, AE, DeepSVDD, and CSI.
	\footnotetext{where all optimizations succeed except the first case in the ablated MLSAD using $A^1$ alone. This may be caused by the non-convex objective.}
	
	\textbf{Results.} 
	 Fig.~\ref{fooling} shows the images found in various horse anomaly detectors. The ablated MLSAD using $A^1$ alone, AE, and DeepSVDD all fail to produce any horse-like images. CSI and the ablated MLSAD using $A^8$ alone indeed produce some images with horse-like patterns. It seems they learned some class discriminative features of the normal data like supervised multi-class classifiers. By contrast, the images found in MLSAD are more recognizable and normal-looking, no matter what images we start from. It indicates that MLSAD tends to unbiasedly detect anomalous features of various levels, and learn better and detailed generative features. As far as we know, deep neural networks widely used today cannot produce such normal-looking images by simply optimizing their outputs~\cite{NNFooled}. Considering that similar results can not be observed in ablated MLSAD detectors, we conclude that the multi-level setting is necessary to build a robust model that probably detects various types of anomalies. 
	\section{Conclusion}
	We first introduced the MRF approximations and argued that images from multi-level MRF approximations can efficiently expose the local abnormal structures of various levels. Then we developed entropy-regularized PatchGAN to generate such images. To fully exploit the local abnormal structures for anomaly detection, we proposed to train multiple level-specific patch-based detectors and calculate the overall anomaly score by aggregating the results over all patches and levels. Diverse multi-level abnormal structures were observed in the generations of entropy-regularized PatchGAN. The results on MNIST, CIFAR10, ImageNet10, and texture datasets show that MLSAD is a general, robust and effective anomaly detection model. They also indicate that unsupervised learning of multi-level visual structures using MRF approximations as contrastive distributions is feasible, which may be extended to broader areas like representation learning. However, MLSAD is computationally expensive for large images due to the multi-level setting and strides of $1\!\times\!1$. In the future, We will investigate the performance of using larger stride sizes, lower resolutions for high-level detectors, and larger level gaps.
	\newpage
	{\small
		\bibliographystyle{ieee_fullname}
		\bibliography{main}

\begin{thebibliography}{10}\itemsep=-1pt

\bibitem{AND}
Davide Abati, Angelo Porrello, Simone Calderara, and Rita Cucchiara.
\newblock Latent space autoregression for novelty detection.
\newblock In {\em CVPR}, pages 481--490, 2019.

\bibitem{GANomaly}
Samet Akcay, Amir Atapour-Abarghouei, and Toby~P Breckon.
\newblock Ganomaly: Semi-supervised anomaly detection via adversarial training.
\newblock In {\em ACCV}, 2018.

\bibitem{VAE}
Jinwon An and Sungzoon Cho.
\newblock Variational autoencoder based anomaly detection using reconstruction
  probability.
\newblock In {\em SNU Data Mining Center, Tech. Rep.}, 2015.

\bibitem{AE2}
Jerone~TA Andrews, Edward~J Morton, and Lewis~D Griffin.
\newblock Detecting anomalous data using auto-encoders.
\newblock {\em International Journal of Machine Learning and Computing},
  6(1):21, 2016.

\bibitem{MINE}
Mohamed~Ishmael Belghazi, Aristide Baratin, Sai Rajeswar, Sherjil Ozair, Yoshua
  Bengio, R.~Devon Hjelm, and Aaron~C. Courville.
\newblock Mutual information neural estimation.
\newblock In {\em ICML}, 2018.

\bibitem{GOAD}
Liron Bergman and Yedid Hoshen.
\newblock Classification-based anomaly detection for general data.
\newblock In {\em ICLR}, 2020.

\bibitem{MVTecAD}
Paul Bergmann, Michael Fauser, David Sattlegger, and Carsten Steger.
\newblock Mvtec ad--a comprehensive real-world dataset for unsupervised anomaly
  detection.
\newblock In {\em CVPR}, 2019.

\bibitem{PSGAN}
Urs Bergmann, Nikolay Jetchev, and Roland Vollgraf.
\newblock Learning texture manifolds with the periodic spatial gan.
\newblock In {\em ICML}, 2017.

\bibitem{AmbientGAN}
Ashish Bora, Eric Price, and Alexandros~G. Dimakis.
\newblock Ambientgan: Generative models from lossy measurements.
\newblock In {\em ICLR}, 2018.

\bibitem{BagNet}
Wieland Brendel and Matthias Bethge.
\newblock Approximating cnns with bag-of-local-features models works
  surprisingly well on imagenet.
\newblock In {\em ICLR}, 2019.

\bibitem{WAIC}
Hyunsun Choi, Eric Jang, and Alexander~A Alemi.
\newblock Waic, but why? generative ensembles for robust anomaly detection.
\newblock {\em arXiv preprint arXiv:1810.01392}, 2018.

\bibitem{ImageNet10}
Jia Deng, W. Dong, R. Socher, L. Li, K. Li, and Li Fei-Fei.
\newblock Imagenet: A large-scale hierarchical image database.
\newblock In {\em CVPR}, 2009.

\bibitem{Geom}
Izhak Golan and Ran El-Yaniv.
\newblock Deep anomaly detection using geometric transformations.
\newblock In {\em NeurIPS}, 2018.

\bibitem{DLbook}
Ian Goodfellow, Yoshua Bengio, and Aaron Courville.
\newblock {\em Deep Learning}.
\newblock MIT Press.
\newblock \url{http://www.deeplearningbook.org}, page 625.

\bibitem{GAN}
Ian~J. Goodfellow, Jean Pouget-Abadie, M. Mirza, Bing Xu, David Warde-Farley,
  Sherjil Ozair, Aaron~C. Courville, and Yoshua Bengio.
\newblock Generative adversarial nets.
\newblock In {\em NIPS}, 2014.

\bibitem{DROCC}
Sachin Goyal, Aditi Raghunathan, Moksh Jain, H. Simhadri, and Prateek Jain.
\newblock Drocc: Deep robust one-class classification.
\newblock In {\em ICML}, 2020.

\bibitem{NCE0}
Michael Gutmann and Aapo Hyv{\"a}rinen.
\newblock Learning features by contrasting natural images with noise.
\newblock In {\em International Conference on Artificial Neural Networks},
  2009.

\bibitem{NCE}
Michael Gutmann and Aapo Hyv{\"a}rinen.
\newblock Noise-contrastive estimation: A new estimation principle for
  unnormalized statistical models.
\newblock In {\em Proceedings of the Thirteenth International Conference on
  Artificial Intelligence and Statistics}, 2010.

\bibitem{ResNet}
Kaiming He, Xiangyu Zhang, Shaoqing Ren, and Jian Sun.
\newblock Deep residual learning for image recognition.
\newblock In {\em CVPR}, 2016.

\bibitem{Relu}
Matthias Hein, Maksym Andriushchenko, and Julian Bitterwolf.
\newblock Why relu networks yield high-confidence predictions far away from the
  training data and how to mitigate the problem.
\newblock In {\em CVPR}, 2018.

\bibitem{softmax}
Dan Hendrycks and Kevin Gimpel.
\newblock A baseline for detecting misclassified and out-of-distribution
  examples in neural networks.
\newblock In {\em ICLR}, 2017.

\bibitem{OE}
Dan Hendrycks, Mantas Mazeika, and Thomas~G. Dietterich.
\newblock Deep anomaly detection with outlier exposure.
\newblock In {\em ICLR}, 2018.

\bibitem{Rot}
Dan Hendrycks, Mantas Mazeika, Saurav Kadavath, and Dawn Song.
\newblock Using self-supervised learning can improve model robustness and
  uncertainty.
\newblock In {\em NeurIPS}, 2019.

\bibitem{pix2pix}
Phillip Isola, Jun-Yan Zhu, Tinghui Zhou, and Alexei~A Efros.
\newblock Image-to-image translation with conditional adversarial networks.
\newblock In {\em CVPR}, 2017.

\bibitem{StyleGAN2}
Tero Karras, Samuli Laine, Miika Aittala, Janne Hellsten, Jaakko Lehtinen, and
  Timo Aila.
\newblock Analyzing and improving the image quality of stylegan.
\newblock In {\em CVPR}, 2020.

\bibitem{app2}
B Kiran, Dilip Thomas, and Ranjith Parakkal.
\newblock An overview of deep learning based methods for unsupervised and
  semi-supervised anomaly detection in videos.
\newblock {\em Journal of Imaging}, 4(2):36, 2018.

\bibitem{GlowFail}
P. Kirichenko, Pavel Izmailov, and A. Wilson.
\newblock Why normalizing flows fail to detect out-of-distribution data.
\newblock In {\em NeurIPS}, 2020.

\bibitem{EBGAN}
Rithesh Kumar, Anirudh Goyal, Aaron~C. Courville, and Yoshua Bengio.
\newblock Maximum entropy generators for energy-based models.
\newblock {\em ArXiv}, abs/1901.08508, 2019.

\bibitem{ConfidentClassifier}
Kimin Lee, Honglak Lee, Kibok Lee, and Jinwoo Shin.
\newblock Training confidence-calibrated classifiers for detecting
  out-of-distribution samples.
\newblock In {\em ICLR}, 2018.

\bibitem{ODIN}
Shiyu Liang, Yixuan Li, and R. Srikant.
\newblock Enhancing the reliability of out-of-distribution image detection in
  neural networks.
\newblock In {\em ICLR}, 2018.

\bibitem{CoordConv}
Rosanne Liu, Joel Lehman, Piero Molino, Felipe~Petroski Such, Eric Frank, Alex
  Sergeev, and Jason Yosinski.
\newblock An intriguing failing of convolutional neural networks and the
  coordconv solution.
\newblock In {\em NeurIPS}, 2018.

\bibitem{FCN}
Jonathan Long, Evan Shelhamer, and Trevor Darrell.
\newblock Fully convolutional networks for semantic segmentation.
\newblock In {\em CVPR}, pages 3431--3440, 2015.

\bibitem{SN}
Takeru Miyato, Toshiki Kataoka, Masanori Koyama, and Yuichi Yoshida.
\newblock Spectral normalization for generative adversarial networks.
\newblock In {\em ICLR}, 2018.

\bibitem{DoGMsKnow}
Eric~T. Nalisnick, Akihiro Matsukawa, Yee~Whye Teh, Dilan G{\"o}r{\"u}r, and
  Balaji Lakshminarayanan.
\newblock Do deep generative models know what they don't know?
\newblock In {\em ICLR}, 2018.

\bibitem{NNFooled}
Anh~M Nguyen, Jason Yosinski, and Jeff Clune.
\newblock Deep neural networks are easily fooled: High confidence predictions
  for unrecognizable images.
\newblock In {\em CVPR}, 2015.

\bibitem{ConAD}
Duc~Tam Nguyen, Zhongyu Lou, Michael Klar, and Thomas Brox.
\newblock Anomaly detection with multiple-hypotheses predictions.
\newblock In {\em ICML}, 2019.

\bibitem{KDE}
E. Parzen.
\newblock On estimation of a probability density function and mode.
\newblock {\em Annals of Mathematical Statistics}, 33:1065--1076, 1962.

\bibitem{OCGAN}
Pramuditha Perera, Ramesh Nallapati, Bing Xiang, and NONE.
\newblock Ocgan: One-class novelty detection using gans with constrained latent
  representations.
\newblock In {\em CVPR}, 2019.

\bibitem{Pidhorskyi2018GenerativePN}
Stanislav Pidhorskyi, Ranya Almohsen, and Gianfranco Doretto.
\newblock Generative probabilistic novelty detection with adversarial
  autoencoders.
\newblock In {\em NeurIPS}, 2018.

\bibitem{GPND}
Stanislav Pidhorskyi, Ranya Almohsen, and Gianfranco Doretto.
\newblock Generative probabilistic novelty detection with adversarial
  autoencoders.
\newblock In {\em NeurIPS}, 2018.

\bibitem{Ratio}
Jie Ren, Peter~J. Liu, Emily Fertig, Jasper Snoek, Ryan Poplin, Mark~A.
  DePristo, Joshua~V. Dillon, and Balaji Lakshminarayanan.
\newblock Likelihood ratios for out-of-distribution detection.
\newblock In {\em NeurIPS}, 2019.

\bibitem{UNet}
Olaf Ronneberger, Philipp Fischer, and Thomas Brox.
\newblock U-net: Convolutional networks for biomedical image segmentation.
\newblock In {\em International Conference on Medical image computing and
  computer-assisted intervention}, 2015.

\bibitem{Review}
Lukas Ruff, J. Kauffmann, Robert~A. Vandermeulen, Gr{\'e}goire Montavon, W.
  Samek, Marius Kloft, Thomas~G. Dietterich, and K. Muller.
\newblock A unifying review of deep and shallow anomaly detection.
\newblock {\em ArXiv}, abs/2009.11732, 2020.

\bibitem{DeepSVDD}
Lukas Ruff, Robert Vandermeulen, Nico Goernitz, Lucas Deecke, Shoaib~Ahmed
  Siddiqui, Alexander Binder, Emmanuel M{\"u}ller, and Marius Kloft.
\newblock Deep one-class classification.
\newblock In {\em ICML}, 2018.

\bibitem{ALOCC}
Mohammad Sabokrou, Mohammad Khalooei, Mahmood Fathy, and Ehsan Adeli.
\newblock Adversarially learned one-class classifier for novelty detection.
\newblock In {\em CVPR}, 2018.

\bibitem{app1}
Daisuke Sato, Shouhei Hanaoka, Yukihiro Nomura, Tomomi Takenaga, Soichiro Miki,
  Takeharu Yoshikawa, Naoto Hayashi, and Osamu Abe.
\newblock A primitive study on unsupervised anomaly detection with an
  autoencoder in emergency head ct volumes.
\newblock In {\em Medical Imaging 2018: Computer-Aided Diagnosis}, 2018.

\bibitem{AnoGAN}
Thomas Schlegl, Philipp Seeb{\"o}ck, Sebastian~M. Waldstein, Ursula~M
  Schmidt-Erfurth, and Georg Langs.
\newblock Unsupervised anomaly detection with generative adversarial networks
  to guide marker discovery.
\newblock In {\em IPMI}, 2017.

\bibitem{OCSVM}
Bernhard Sch{\"o}lkopf, John~C Platt, John Shawe-Taylor, Alex~J Smola, and
  Robert~C Williamson.
\newblock Estimating the support of a high-dimensional distribution.
\newblock {\em Neural computation}, 13(7):1443--1471, 2001.

\bibitem{FirstRobust}
Lukas Schott, Jonas Rauber, M. Bethge, and W. Brendel.
\newblock Towards the first adversarially robust neural network model on mnist.
\newblock In {\em ICLR}, 2019.

\bibitem{complexity}
Joan Serr{\`a}, David {\'A}lvarez, Vicenç G{\'o}mez, Olga Slizovskaia,
  Jos{\'e}~F. N{\'u}{\~n}ez, and Jordi Luque.
\newblock Input complexity and out-of-distribution detection with
  likelihood-based generative models.
\newblock In {\em ICLR}, 2020.

\bibitem{LessBiased}
Alireza Shafaei, Mark Schmidt, and James~J. Little.
\newblock A less biased evaluation of out-of-distribution sample detectors.
\newblock In {\em BMVC}, 2019.

\bibitem{SinGAN}
Tamar~Rott Shaham, Tali Dekel, and Tomer Michaeli.
\newblock Singan: Learning a generative model from a single natural image.
\newblock In {\em ICCV}, 2019.

\bibitem{CSI}
Jihoon Tack, Sangwoo Mo, Jongheon Jeong, and Jinwoo Shin.
\newblock Csi: Novelty detection via contrastive learning on distributionally
  shifted instances.
\newblock In {\em NeurIPS}, 2020.

\bibitem{OODGen}
Sachin Vernekar, Ashish Gaurav, Vahdat Abdelzad, Taylor Denouden, Rick Salay,
  and Krzysztof Czarnecki.
\newblock Out-of-distribution detection in classifiers via generation.
\newblock {\em NeurIPS workshop}, 2019.

\bibitem{Pooling}
Yun Wang, Juncheng Li, and Florian Metze.
\newblock A comparison of five multiple instance learning pooling functions for
  sound event detection with weak labeling.
\newblock In {\em ICASSP 2019-2019 IEEE International Conference on Acoustics,
  Speech and Signal Processing (ICASSP)}, 2019.

\bibitem{PatchSVDD}
Jihun Yi and S. Yoon.
\newblock Patch svdd: Patch-level svdd for anomaly detection and segmentation.
\newblock In {\em ACCV}, 2020.

\bibitem{ASG}
Yang Yu, Wei-Yang Qu, Nan Li, and Zimin Guo.
\newblock Open-category classification by adversarial sample generation.
\newblock In {\em IJCAI}, 2017.

\bibitem{PT}
Junbo Zhao, Michael Mathieu, and Yann LeCun.
\newblock Energy-based generative adversarial networks.
\newblock In {\em ICLR}, 2017.

\bibitem{ComplentaryGAN}
Panpan Zheng, Shuhan Yuan, Xintao Wu, Jun~Yu Li, and Aidong Lu.
\newblock One-class adversarial nets for fraud detection.
\newblock In {\em AAAI}, 2018.

\end{thebibliography}
	}
	\newpage
	\onecolumn
	\setcounter{theorem}{0}
	\addcontentsline{toc}{section}{Appendix}
    \renewcommand{\thesubsection}{\Alph{subsection}}
	\section{Appendix}
    \subsection{Proofs}
    \begin{theorem}
    	\label{theorem_supp_eq_app}
    	Let $S_w=\{\bm{x}|\forall c\in C^w,p_{\bm{X}_c}(\bm{x}_c)>0\}$, then $supp(q^w_{\bm{X}})=S_w$. In other words, the support set of $q^w_{\bm{X}}$ is identical to the set of all images where each patch of $w\times w$ pixels is normal.
    \end{theorem}
    \textit{Proof.} Firstly, we prove $supp(q^w_{\bm{X}})\subseteq S_w$. \\For all $\bm{x} \in supp(q^w_{\bm{X}})$, we have $q^w_{\bm{X}}(\bm{x})>0$. Let ${\bar{c}}$ be the pixels outside the patch $c$. Since 
    \begin{equation*}
    q^w_{\bm{X}}(\bm{x})= q_{\bm{X}_c}^w(\bm{x}_c)q_{\bm{X}_{\bar{c}}|\bm{X}_c}^w(\bm{x}_{\bar{c}}|\bm{x}_c),
    \end{equation*} 
    we have $q_{\bm{X}_c}^w(\bm{x}_c)>0$. By the definition of MRF approximations (Definition 1), we know that $\forall c\in C^w$, $q_{\bm{X}_c}^w(\bm{x}_c)=p_{\bm{X}_c}(\bm{x}_c)$. Therefore, $\forall \bm{x} \in supp(q^w_{\bm{X}}), p_{\bm{X}_c}(\bm{x}_c)>0$ holds for all $c\in C^w$. Because $S_w=\{\bm{x}|\forall c\in C^w,p_{\bm{X}_c}(\bm{x}_c)>0\}$, for all $\bm{x}\in supp(q^w_{\bm{X}})$ we have $\bm{x}\in S_w$. It follows that $supp(q^w_{\bm{X}})\subseteq S_w$.\\
    Then, we prove $S_w\subseteq supp(q^w_{\bm{X}})$.\\
    By the definition of MRF approximations, we know that the $w$th-level MRF approximation can be written as
    \begin{equation}
    \label{factor}
    q^w_{\bm{X}}(\bm{x})=\prod_{c\in C^w} \phi_c(\bm{x}_c).
    \end{equation}
    Since
    \begin{equation*}
    q_{\bm{X}_c}^w(\bm{x}_c)=\int_{\bm{x}_{\bar{c}}}q_{\bm{X}}^w(\bm{x}_c,\bm{x}_{\bar{c}})d\bm{x}_{\bar{c}},
    \end{equation*}
    we have
    \begin{equation}
    \label{factor_int}
    \begin{aligned}
    q_{\bm{X}_c}^w(\bm{x}_c)=&\int_{\bm{x}_{\bar{c}}}\prod_{c'\in C^w} \phi_{c'}(\bm{x}_{c'})d\bm{x}_{\bar{c}}\\
     =&\phi_{c}(\bm{x}_{c})\int_{\bm{x}_{\bar{c}}}\prod_{c'\in C^w\setminus\{c\}} \phi_{c'}(\bm{x}_{c'})d\bm{x}_{\bar{c}}\\
     =&\phi_{c}(\bm{x}_{c})\psi_c(\bm{x}_c),
    \end{aligned}
    \end{equation}
    where $\psi_c(\bm{x}_c):=\int_{\bm{x}_{\bar{c}}}\prod_{c'\in C^w\setminus\{c\}} \phi_{c'}(\bm{x}_{c'})d\bm{x}_{\bar{c}}$. Then for all $\bm{x}$ such that $\forall c\in C^w,p_{\bm{X}_c}(\bm{x}_c)>0$, since $q_{\bm{X}_c}^w(\bm{x}_c)=p_{\bm{X}_c}(\bm{x}_c)$, we have $q_{\bm{X}_c}^w(\bm{x}_c)>0$. Then using Equation~\ref{factor} and Equation~\ref{factor_int}, we can get $\forall c\in C^w,\phi_{c}(\bm{x}_{c})>0$ and $q^w_{\bm{X}}(\bm{x})>0$. Hence $\forall \bm{x} \in S_w, \bm{x} \in supp(q^w_{\bm{X}})$ holds and thus $S_w\subseteq supp(q^w_{\bm{X}})$.\\
    Finally, we conclude that $supp(q^w_{\bm{X}})=S_w$. 
    
    \begin{theorem}
    	\label{theorem_max_entropy_app}
	The maximum entropy distribution satisfying $\forall c\!\in\!C^w, q_{\bm{X}_c}(\bm{x}_c)\!=\! p_{\bm{X}_c}(\bm{x}_c)$ can be factored as $\prod_{c\in C^w} \phi_c(\bm{x}_c)$.
    \end{theorem}
    \textit{Proof.} We prove it with the calculus of variations and Lagrange multipliers. Firstly, we can have the functional constraints:
    \begin{equation*}
    G_{c}[q_{\bm{X}}]:= q_{\bm{X}_c}(\bm{x}_{c})-p_{\bm{X}_{c}}(\bm{x}_{c})=0,\forall c\in C^w.
    \end{equation*}
    Additionally, we define a functional constraint that our density must sum to $1$:
    \begin{equation*}
    G_0[q_{\bm{X}}]:=\int_{\bm{x}} q_{\bm{X}}(\bm{x}) d\bm{x}-1=0.
    \end{equation*}
    Let us use $H$ to denote entropy, then the differential entropy functional can be written as
    \begin{equation*}
    H[q_{\bm{X}}]:=-\int_{\bm{x}} q_{\bm{X}}(\bm{x})\log{q_{\bm{X}}(\bm{x})} d\bm{x}.
    \end{equation*}
    To maximize entropy $H$ subject to $G_0$ and a serial of patch constraints $\{G_c|c\in C^w\}$, we put them together and consider the functional:
    \begin{equation*}
    \begin{aligned}
    &J[q_{\bm{X}}]=\int_{\bm{x}} -q_{\bm{X}}(\bm{x})\log q_{\bm{X}}(\bm{x})d\bm{x}-\\
    &\lambda_0 [\int_{\bm{x}} q_{\bm{X}}(\bm{x})d\bm{x}-1]-\\
    &\sum_{c}\int_{\bm{x}_{c}}\lambda_{c}(\bm{x}_{c}) [\int_{\bm{x}_{\bar{c}}}q_{\bm{X}}(\bm{x})d\bm{x}_{\bar{c}}-p_{\bm{X}_{c}}(\bm{x}_{c})]d\bm{x}_{c},\\
    \end{aligned}
    \end{equation*}
    where $\lambda_0$ and $\lambda_c$ are Lagrange multipliers.
    The final term can be arranged into 
    \begin{equation*}
    \sum_{c}\int_{\bm{x}}\lambda_{c}(\bm{x}_{c}) q_{\bm{X}}(\bm{x})d\bm{x}-\int_{\bm{x}_{c}}\lambda_{c}(\bm{x}_{c})p_{\bm{X}_{c}}(\bm{x}_{c})d\bm{x}_{c}.
    \end{equation*}
    The entropy attains an extremum when the functional derivative is equal to zero:
    \begin{equation*}
    \frac{\delta{J}}{\delta{q_{\bm{X}}}}=-1-\log{q_{\bm{X}}(\bm{x})}-\lambda_0-\sum_{c}\lambda_{c}(\bm{x}_{c})=0.
    \end{equation*}
    We can get 
    \begin{equation*}
    \lambda_0 = -1,
    q_{\bm{X}}(\bm{x})=e^{-\sum_c\lambda_c(\bm{x}_c)}.
    \end{equation*}
    Therefore the maximum entropy distribution $q_{\bm{X}}(\bm{x})$ can be factored as $\prod_c \phi_c(\bm{x}_c)$.
    \subsection{Discussion on the position anomaly pair}
\begin{figure}[!htbp]
    \begin{minipage}{0.58\linewidth}
    	\centering
    	\includegraphics[width=\linewidth]{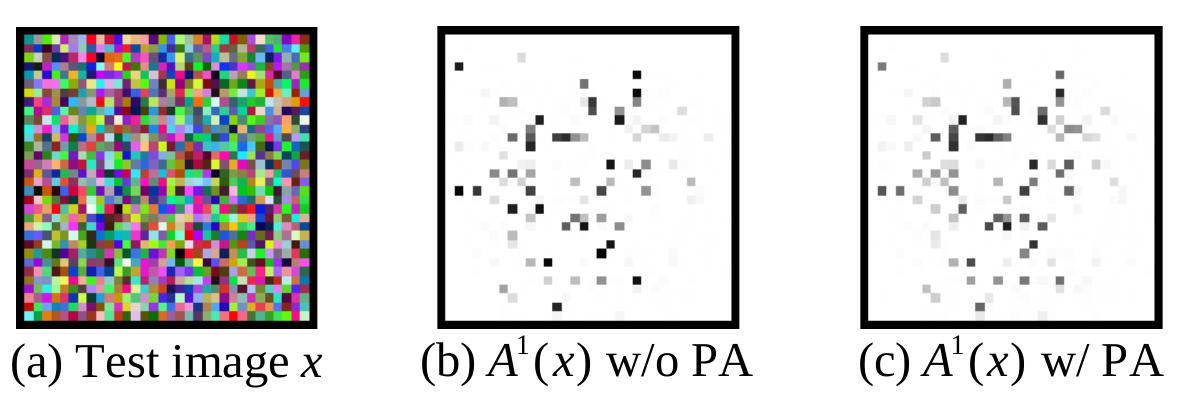}
    	\caption{Both the pixel-level detectors trained with and without the position anomaly pair can detect out almost all the colorful pixels.}
    	\label{fig:noise_score}
    \end{minipage}
    \begin{minipage}{0.42\linewidth}
    	\centering
    	\includegraphics[width=\linewidth]{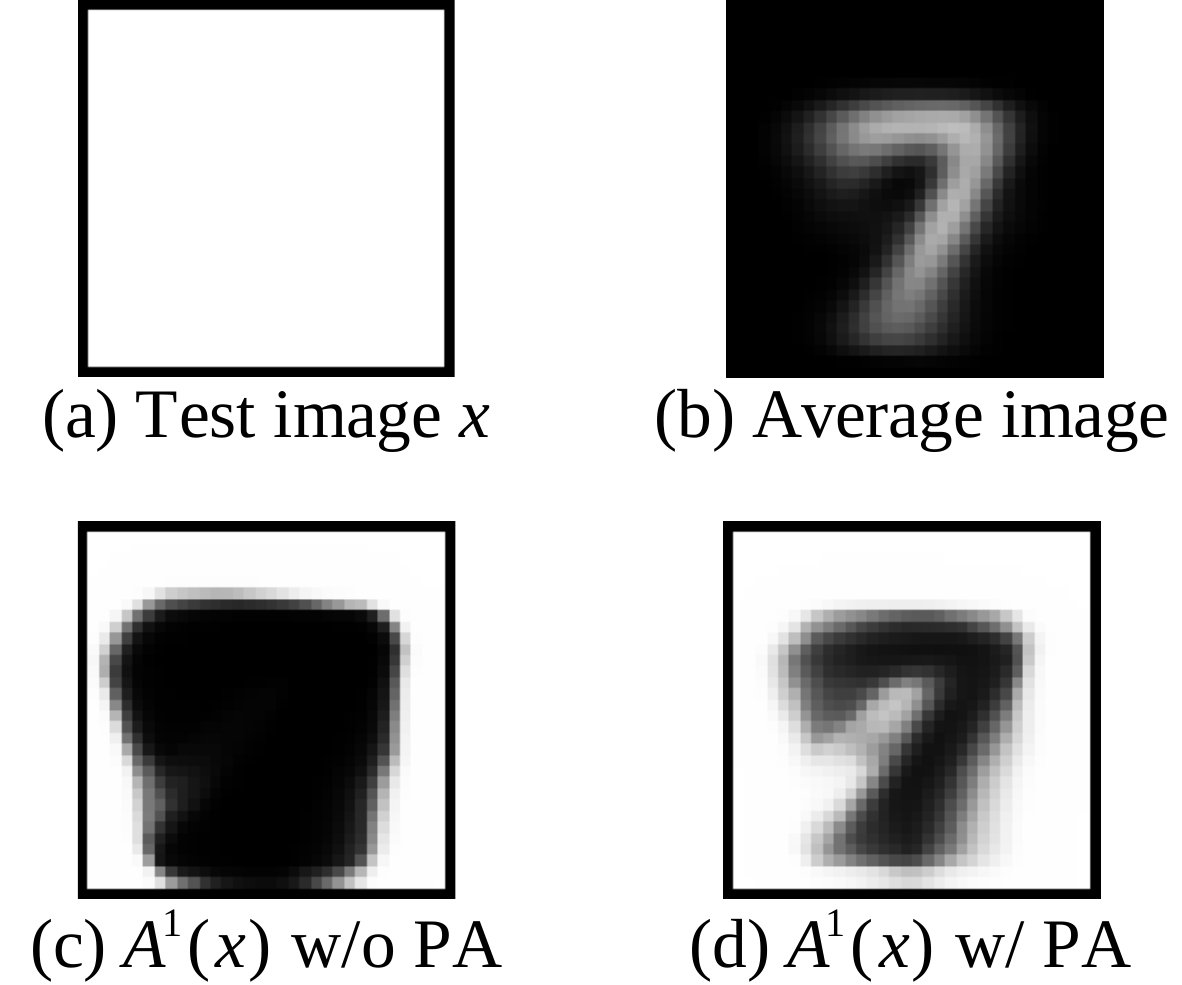}
    	\caption{The position anomaly pair helps the pixel-level detector learn a better position-aware anomaly score.}
    	\label{fig:PA score}
    \end{minipage}
    \end{figure}

    We train a pixel-level detector $A^1$ and an ablated $A^1$ by removing the position anomaly pair (denoted by w/o PA) on images of MNIST digit 7, and compare them by visualizing their score maps. We first test both detectors on a noise image. The results are shown in Figure~\ref{fig:noise_score}. We find that both the detectors can output high anomaly scores for almost all the colorful pixels. We then test both detectors on an image whose all pixels are white, as shown in Figure~\ref{fig:PA score}. Since the border regions of digit 7 images are always black, a perfect pixel-level anomaly detector should output high anomaly scores for such regions. It is observed that both the detectors (trained with and without the position anomaly pair) work nicely to a certain extent. However, we find that the detector trained with the position anomaly pair (w/ PA) outputs a score map with a more concrete shape that looks like the average image. Considering that the pixel values in MNIST images are almost binary, the pixel brightness in the average image can be treated as the probability that the corresponding pixel is white, and the observations in Figure~\ref{fig:noise_score}, we conclude that the ablated pixel-level detector (w/o PA) focus more on learning the common color statistics shared in various locations, and the position anomaly pair indeed helps the pixel-level detector to learn the position-dependent abnormal cues.
    
    \subsection{Experimental details}
    \subsubsection{Maximum entropy-regularized PatchGAN.}
    We use Pytorch library to build and train our models. The intensity range of the images is normalized into $[-1,1]$. 
    As StyleGAN2~\cite{StyleGAN2}, we use minibatch standard deviation layers in the discriminators, and use bilinear upsampling layers in the generators. The generators are based on the UNet~\cite{UNet} architecture. Both SN and batch normalization are used at all non-output layers. For MNIST, CIFAR10 and ImageNet10 dataset, we generate images with a size of $32\times32$, the network detail is shown in Table~\ref{g}. We use a similar but deeper network with a ouput size of $256\times256$ for textures in MVTec AD. All generators produce images from a noise $z$ with the same shape to the training images. For the discriminators, we use spectral normalization (SN)~\cite{SN} at all layers, the network details are shown in Table~\ref{d2},~\ref{d4}, and~\ref{d8}. The $\beta$ used for training generators (in Equation $2$ of the main paper) is dynamically calculated to clip the gradient of the entropy term~\cite{MINE} so that it will not overwhelm the local structure term:
    \begin{equation}
        \beta = \min(\frac{\|\bm{g}_{adv}\|}{\|\bm{g}_I\|+10^{-8}},1)
    \end{equation}
    where $\bm{g}_{adv}$ is the gradient of $\mathcal{L}_{adv}$ respect to $G$, and $\bm{g}_I$ is the gradient from $\mathcal{I}_{\Theta}$. The statistics network $T$ takes the concatenation of $z$ and $G(z)$ as input and uses a multi-scale pooling operation to calculate the final output, which is expected to be better for producing diverse recombinations. We train them with the Adam optimizer, a batch size of 32, and a max epoch of 300. The optimizer settings are shown in Table~\ref{tab:optim}:
    \begin{table}[!htbp]
    	\begin{minipage}{.5\linewidth}
    		\centering
    		\caption{Discriminator network for $2$th-level MRF approximation.}
    		\resizebox{0.98\textwidth}{!}{
    			\begin{tabular}{l|l|l|l|l|l}
    				\toprule
    				Layer& Output&	Kernel&	Stride&	Activation&	Normalization\\
    				& shape& size&	& function& function\\
    				\midrule
    				Input $\bm{x}$,$\bm{r}$&	32$\times$32$\times$5&	 &	 &	  &	 \\
    				Convolution&31$\times$31$\times$64&2$\times$2 & 1$\times$1 & LReLU&SN\\
    				Minibatch stddev &31$\times$31$\times$68&-&-&-&-\\
    				Convolution&31$\times$31$\times$128&1$\times$1 & 1$\times$1 & LReLU&SN\\
    				ResBlock(128)&31$\times$31$\times$128&-&-&-&-\\
    				ResBlock(128)&31$\times$31$\times$128&-&-&-&-\\
    				Convolution&31$\times$31$\times$1&1$\times$1 & 1$\times$1 & Sigmoid&None\\
    				\bottomrule
    		\end{tabular}}
    		\label{d2}
    	\end{minipage}
    	\begin{minipage}{.5\linewidth}
    		\centering
    		\caption{Discriminator network for $4$th-level MRF approximation.}
    		\resizebox{0.98\textwidth}{!}{
    			\begin{tabular}{l|l|l|l|l|l}
    				\toprule
    				Layer& Output&	Kernel&	Stride&	Activation&	Normalization\\
    				& shape& size&	& function& function\\
    				\midrule
    				Input $\bm{x}$,$\bm{r}$&	32$\times$32$\times$5&	 &	 &	  &	 \\
    				Convolution&29$\times$29$\times$128&4$\times$4 & 1$\times$1 & LReLU&SN\\
    				Minibatch stddev &29$\times$29$\times$132&-&-&-&-\\
    				Convolution&29$\times$29$\times$256&1$\times$1 & 1$\times$1 & LReLU&SN\\
    				ResBlock(256)&29$\times$29$\times$256&-&-&-&-\\
    				ResBlock(256)&29$\times$29$\times$256&-&-&-&-\\
    				Convolution&29$\times$29$\times$1&1$\times$1 & 1$\times$1 & Sigmoid&None\\
    				\bottomrule
    		\end{tabular}}
    		\label{d4}
    	\end{minipage}
    \end{table}
    \begin{table}
    	\begin{minipage}{.6\linewidth}
    		\caption{Discriminator network for $8$th-level MRF approximation.}
    		\centering
    		\resizebox{0.98\textwidth}{!}{
    			\begin{tabular}{l|l|l|l|l|l}
    				\toprule
    				Layer&	Output &	Kernel &	Stride&	Activation &	Normalization \\
    				&	 shape&	 size&	size&	 function&	 function\\
    				\midrule
    				Input $\bm{x}$,$\bm{r}$&	32$\times$32$\times$5&	 &	 &	  &	 \\
    				Convolution&29$\times$29$\times$64&4$\times$4 & 1$\times$1 & LReLU&SN\\
    				Minibatch stddev &29$\times$29$\times$68&-&-&-&-\\
    				Convolution&27$\times$27$\times$128&3$\times$3 & 1$\times$1 & LReLU&SN\\
    				Minibatch stddev &27$\times$27$\times$132&-&-&-&-\\
    				Convolution&25$\times$25$\times$256&3$\times$3 & 1$\times$1 & LReLU&SN\\
    				Minibatch stddev &25$\times$25$\times$260&-&-&-&-\\
    				Convolution&25$\times$25$\times$512&1$\times$1 & 1$\times$1 & LReLU&SN\\
    				ResBlock(256)&25$\times$25$\times$512&-&-&-&-\\
    				ResBlock(256)&25$\times$25$\times$512&-&-&-&-\\
    				Convolution&25$\times$25$\times$1&1$\times$1 & 1$\times$1 & Sigmoid&None\\
    				\bottomrule
    			\end{tabular}
    			\label{d8}}
    	\end{minipage}
    	\hfill
    	\begin{minipage}{.4\linewidth}
    		\caption{Generator network for every level}
    		\centering
    		\resizebox{0.98\textwidth}{!}{
    			\begin{tabular}{l|l|l|l|l|l}
    				\toprule
    				Layer & Output  & Kernel &	Stride&Activation &	Normalization \\
    				&  shape &  size&	size& function&	 function\\
    				\midrule
    				Input $\bm{z}$ & 32$\times$32$\times$3 &	 &	  &	 \\
    				Convolution & 16$\times$16$\times$64&4$\times$4 & 2$\times$2 & LReLU&BN\\
    				\midrule
    				Convolution $c_1$&16$\times$16$\times$64& 1$\times$1 & 1$\times$1 & LReLU&BN\\
    				\midrule
    				Convolution &8$\times$8$\times$128& 4$\times$4 & 2$\times$2 & LReLU&BN\\
    				Convolution $c_2$&8$\times$8$\times$128& 1$\times$1 & 1$\times$1 & LReLU&BN\\
    				\midrule
    				Convolution &4$\times$4$\times$256& 4$\times$4 & 2$\times$2 & LReLU&BN\\
    				Convolution $c_3$&4$\times$4$\times$256& 1$\times$1 & 1$\times$1 & LReLU&BN\\
    				\midrule
    				Convolution &1$\times$1$\times$512& 4$\times$4 & 1$\times$1 & LReLU&BN\\
    				\midrule
    				FC \& reshape &4$\times$4$\times$256& - & - & LReLU&BN\\
    				\midrule
    				Concat($\cdot,c_3$)&4$\times$4$\times$512 & - & -&-&-\\
    				Upsample&$8\times$8$\times$512& - & - & -&-\\
    				Convolution &8$\times$8$\times$128& 3$\times$3 & 1$\times$1 & LReLU&BN\\
    				Convolution &8$\times$8$\times$128& 1$\times$1 & 1$\times$1 & LReLU&BN\\
    				\midrule
    				Concat($\cdot,c_2$)&8$\times$8$\times$256 & - & -&-&-\\
    				Upsample&$16\times$16$\times$256& - & - & -&-\\
    				Convolution &16$\times$16$\times$64& 3$\times$3 & 1$\times$1 & LReLU&BN\\
    				Convolution &16$\times$16$\times$64& 1$\times$1 & 1$\times$1 & LReLU&BN\\
    				\midrule
    				Concat($\cdot,c_1$)&16$\times$16$\times$128 & - & -&-&-\\
    				Upsample&$32\times$32$\times$128& - & - & -&-\\
    				Convolution &32$\times$32$\times$64& 3$\times$3 & 1$\times$1 & LReLU&BN\\
    				Convolution &32$\times$32$\times$3& 1$\times$1 & 1$\times$1 & Tanh&None\\
    				\bottomrule
    		\end{tabular}}
    		\label{g}
    	\end{minipage}
    \end{table}
    
    \begin{table}[!h]
    \begin{minipage}{0.5\textwidth}
    	\centering
    	\caption{Optimizer settings for training GANs.}
    	\begin{tabular}{l|l|l|l}
    		\toprule
    		Network & Learning rate&$\beta_1$ & $\beta_2$\\
    		\midrule
    		Discriminators&	$4\times10^{-4}$	 &	$0$ &	$0.9$ \\
    		Generators&$1\times10^{-4}$ & $0$ &	$0.9$\\
    		Statistics networks&$4\times10^{-4}$ & $0$ &	$0.9$\\
    		\bottomrule
        \end{tabular}
    \label{tab:optim}
    \end{minipage}
    \begin{minipage}{0.5\textwidth}
        \centering
     	\caption{Hyper-parameter search ranges for training detectors.}
     	\begin{tabular}{l|l}
     		\toprule
     		hyperparameter&range\\
     		\midrule
     		batch size&\{8, 16, 32, 64\}\\
     		weight decay&\{1E-3, 5E-4, 1E-4, 5E-5\}\\
     		noise&\{0, 0.05 , 0.1\}\\
     		$\alpha_1$&\{0.4, 0.6, 1.0\}\\
     		$\alpha_2$&\{0.4, 0.6, 1.0\}\\
     		num\_block&\{1, 2\}\\
     		\bottomrule
     	\end{tabular}
     	\label{tab:search_range}
    \end{minipage}
    \end{table}
    \subsubsection{Details of level-specific detectors.} Level-specific detectors use no normalization operation for all. We add a Gaussian noise with a variance of $0.05$ for the input layer, and use Dropout operations in the ResBlock layers~\cite{ResNet}. They are expected to avoid overfitting. We use Adam as the optimizer with a initial learning rate of $10^{-3}$, $\beta_1=0.9$, and $\beta_2=0.999$. For both MNIST and textures in MVTec AD, $num\_block=1, weight\_decay=10^{-3}$, and the learning rates decay by a factor of 0.3 at 50th and 75th epoch. For both CIFAR10 and ImageNet10, $num\_block=2,weight\_decay=10^{-4}$, and the learning rates decay by a factor of at 100th and 200th epoch. We train level-specific detectors for a total of 100 epochs on MNIST and textures, and 250 epochs on CIFAR10 and ImageNet10, all with a batch size of 8. The hyper-parameters $\alpha_1$ and $\alpha_2$ are set to $0.6$ and $0.4$ for all levels and all datasets. We perform a non-exaustive search for the hyper-parameters in the ranges shown in Table~\ref{tab:search_range}.
    
    \begin{figure}[!htbp]
    	\centering
    	\includegraphics[width=0.8\linewidth]{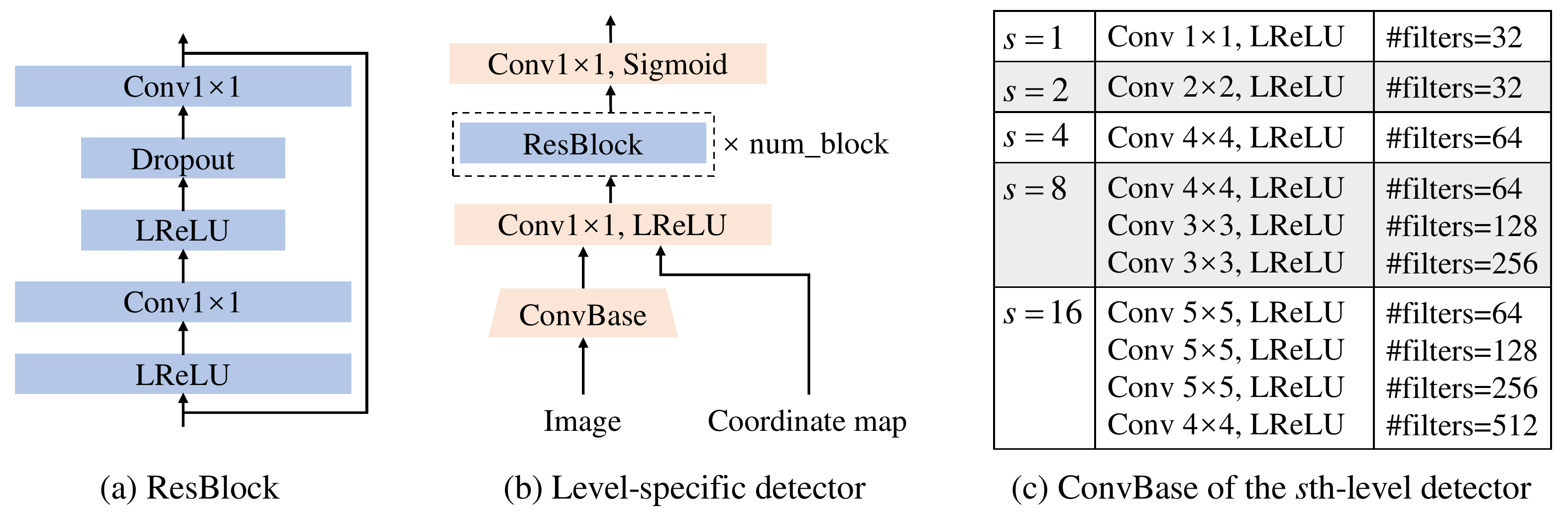}
    	\caption{The networks of level-specific detectors.}
    	\label{fig:detectors}
    \end{figure}
    \subsubsection{Details of investigating the robustness.} The total objective for the fooling test is to find $N$ images $\bm{X}_{\bm{B}}=\{\bm{x}_1,\bm{x}_2,...,\bm{x}_N\}$ that minimize $\frac{1}{N}\sum_i A(\bm{x}_i) + \lambda L_{PT}$. Where $\lambda=10^{-5}$ is used in our experiments. We initialize $\bm{x}_i$ with images sampled from various datasets. Then we update them using the Adam optimizer with the learning rate of $0.02$, $\beta_1=0.5$ and $\beta_2=0.9$. The optimization will always be constrained in the image space by the clipping operation after each update: $\bm{x}_i$ = clip($\bm{x}_i$, -1.0, 1.0) or clip($\bm{x}_i$, 0.0, 1.0) (depending on the input range of the tested model).
    \subsection{More Results on Textures}
    \begin{figure}[!htbp]
    	\centering
    	\includegraphics[width=0.72\linewidth]{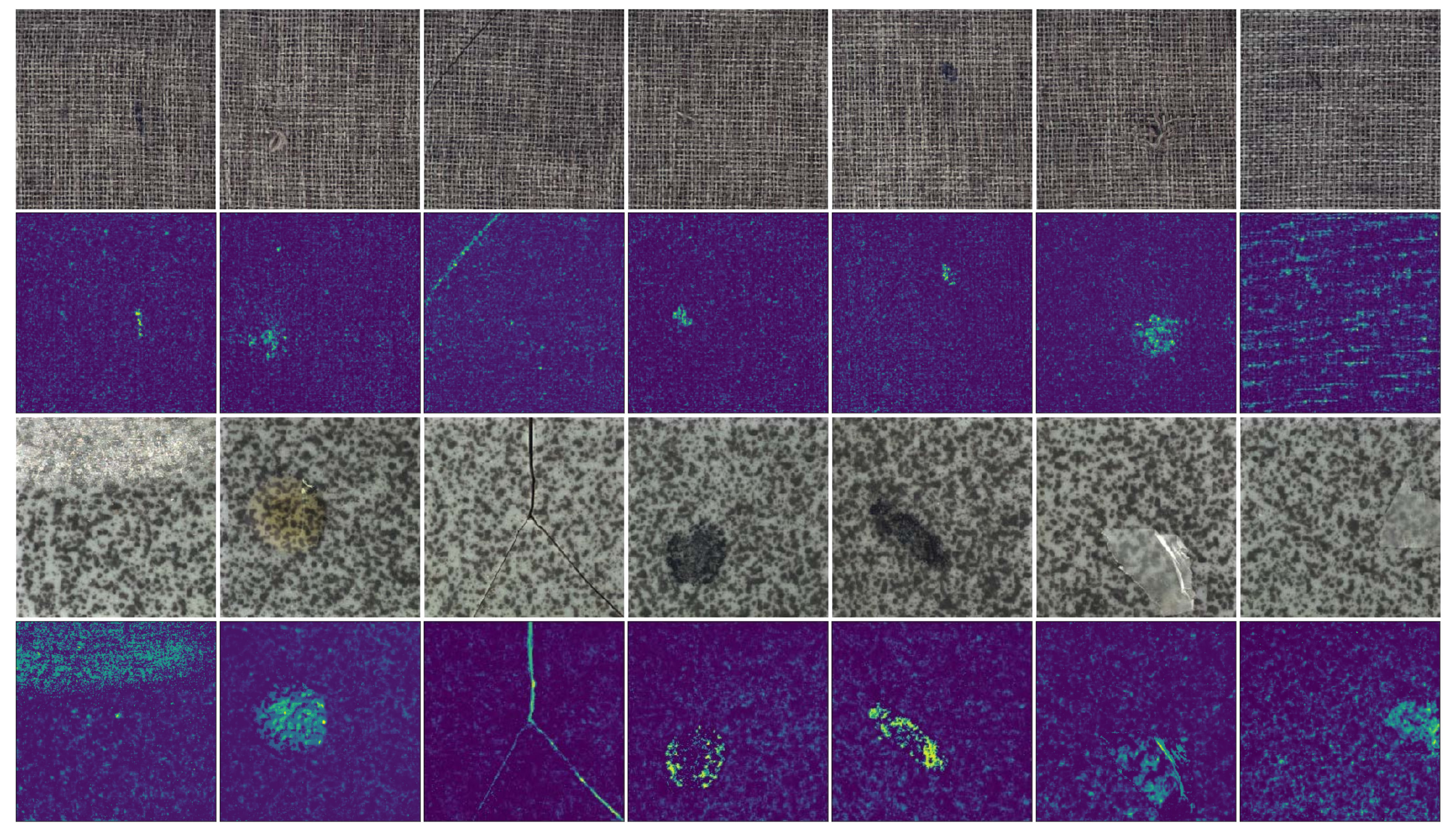}
    	\caption{Qualitative results on the textures in MVTec AD. The line 1 and line 3 are the test anomalous images of 'carpet' category and 'tile' category, respectively. The line 2 and line 4 are the corresponding (aggregated) pixel-level anomaly scores. We remark that the results can be further improved by using softmax pooling for the multi-level aggregation.}
    	\label{texture2}
    \end{figure}
    We present the additional visualizations of the pixel-level detection results (using average pooling for the multi-level aggregation) on textures of MVTec AD in Fig.~\ref{texture2}. 
    
    \subsection{More results on robustness}
\begin{figure}[!htbp]
    	\centering
    	\includegraphics[width=0.8\linewidth]{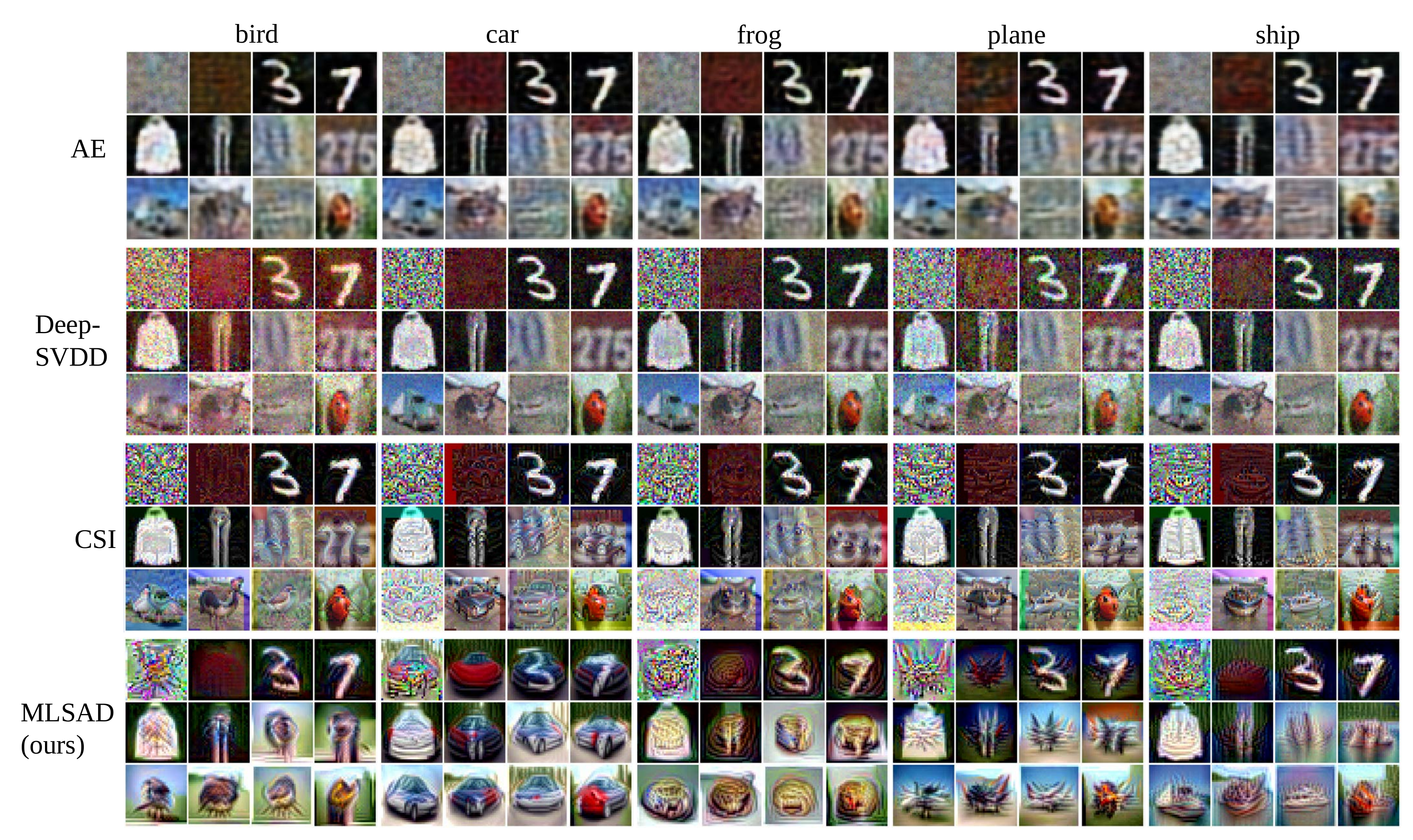}
    	\caption{Images found with lower anomaly scores than 95\% test normal images' via gradient decent. The images found in MLSAD have the smallest perceptual differences from the normal classes.}
    	\label{am2}
    \end{figure}

\begin{figure}[!htbp]
    	\centering
    	\includegraphics[width=0.8\linewidth]{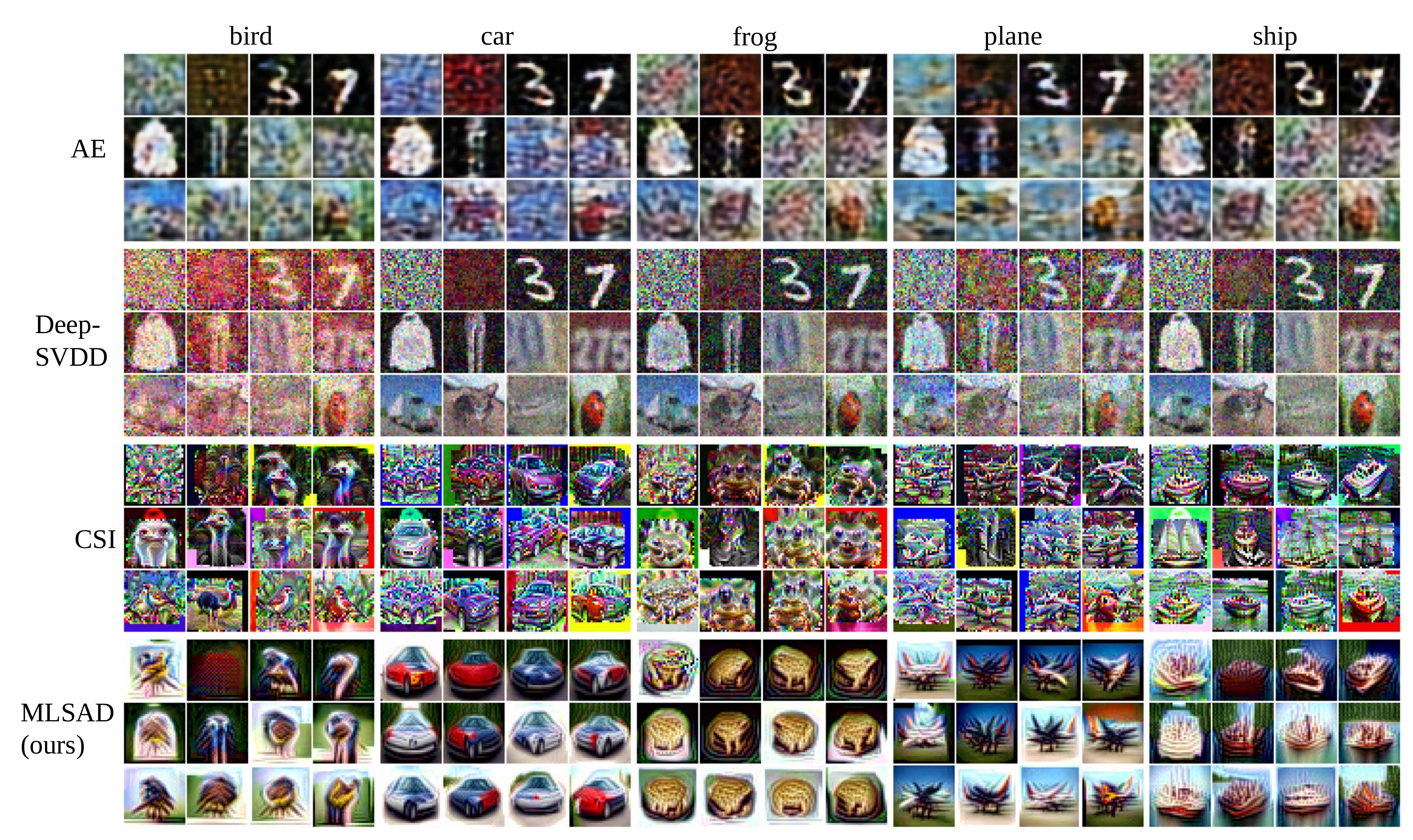}
    	\caption{Images found with much more gradient decent steps (until the images have almost no changes). We again find that MLSAD produces the most normal-looking and recognizable images.}
    	\label{am3}
    \end{figure}

    We present the additional results on the robustness in Fig.~\ref{am2} and Fig.~\ref{am3}, the detectors tesetd here are trained on CIFAR10 classes 'bird', 'plane', 'frog', 'car' and 'ship'. We compare the MLSAD with AE, DeepSVDD and CSI. Results demonstrate the MLSAD can learn the more detailed and complete statistics of the normal data.
    \vspace{-0.4cm}
    \vspace{-0.4cm}

\end{document}